\documentclass[twoside,twocolumn,9pt]{article}
\usepackage{extsizes}
\usepackage{ mathrsfs }
\usepackage[super,sort&compress,comma]{natbib} 
\usepackage[version=3]{mhchem}
\usepackage[left=1.5cm, right=1.5cm, top=1.785cm, bottom=2.0cm]{geometry}
\usepackage{balance}
\usepackage{mathptmx}
\usepackage{amsmath,amssymb}
\usepackage{bm}
\usepackage{sectsty}
\usepackage{graphicx} 
\usepackage{lastpage}
\usepackage[format=plain,justification=justified,singlelinecheck=false,font={stretch=1.125,small,sf},labelfont=bf,labelsep=space]{caption}
\usepackage{float}
\usepackage{fancyhdr}
\usepackage{fnpos}
\usepackage[english]{babel}
\addto{\captionsenglish}{%
  
}
\usepackage{array}
\usepackage{droidsans}
\usepackage{charter}
\usepackage[T1]{fontenc}
\usepackage[usenames,dvipsnames]{xcolor}
\usepackage{setspace}
\usepackage[compact]{titlesec}
\usepackage{hyperref}

\usepackage{color}
\newcommand{\exciting}{{\usefont{T1}{lmtt}{b}{n}exciting}}
\newcommand{\excitingtools}{{\usefont{T1}{lmtt}{b}{n}excitingtools}}

\newcommand{\eg}{{\it e.g.}, }
\newcommand{\ie}{{\it i.e.}, }

\usepackage{epstopdf}

\definecolor{cream}{RGB}{222,217,201}

\begin{document}

\pagestyle{fancy}
\thispagestyle{plain}
\fancypagestyle{plain}{
\renewcommand{\headrulewidth}{0pt}
}

\makeFNbottom
\makeatletter
\renewcommand\LARGE{\@setfontsize\LARGE{15pt}{17}}
\renewcommand\Large{\@setfontsize\Large{12pt}{14}}
\renewcommand\large{\@setfontsize\large{10pt}{12}}
\renewcommand\footnotesize{\@setfontsize\footnotesize{7pt}{10}}
\makeatother

\renewcommand{\thefootnote}{\fnsymbol{footnote}}
\renewcommand\footnoterule{\vspace*{1pt}%
\color{cream}\hrule width 3.5in height 0.4pt \color{black}\vspace*{5pt}} 
\setcounter{secnumdepth}{5}

\makeatletter 
\renewcommand\@biblabel[1]{#1}            
\renewcommand\@makefntext[1]%
{\noindent\makebox[0pt][r]{\@thefnmark\,}#1}
\makeatother 
\renewcommand{\figurename}{\small{Fig.}~}
\sectionfont{\sffamily\Large}
\subsectionfont{\normalsize}
\subsubsectionfont{\bf}
\setstretch{1.125} 
\setlength{\skip\footins}{0.8cm}
\setlength{\footnotesep}{0.25cm}
\setlength{\jot}{10pt}
\titlespacing*{\section}{0pt}{4pt}{4pt}
\titlespacing*{\subsection}{0pt}{15pt}{1pt}

\fancyfoot{}
\fancyfoot[LE]{\footnotesize{\sffamily{\thepage~\textbar\hspace{3.45cm}}}}
\fancyhead{}
\renewcommand{\headrulewidth}{0pt} 
\renewcommand{\footrulewidth}{0pt}
\setlength{\arrayrulewidth}{1pt}
\setlength{\columnsep}{6.5mm}
\setlength\bibsep{1pt}

\makeatletter 
\newlength{\figrulesep} 
\setlength{\figrulesep}{0.5\textfloatsep} 

\newcommand{\topfigrule}{\vspace*{-1pt}%
\noindent{\color{cream}\rule[-\figrulesep]{\columnwidth}{1.5pt}} }

\newcommand{\botfigrule}{\vspace*{-2pt}%
\noindent{\color{cream}\rule[\figrulesep]{\columnwidth}{1.5pt}} }

\newcommand{\dblfigrule}{\vspace*{-1pt}%
\noindent{\color{cream}\rule[-\figrulesep]{\textwidth}{1.5pt}} }

\makeatother

\twocolumn[
\begin{@twocolumnfalse}
\vspace{1em}
\sffamily
\begin{tabular}{m{15.0cm} p{1cm} }

\noindent\LARGE{\textbf{How big is Big Data?}} 
\vspace{0.3cm} \vspace{0.3cm} \\
\noindent\large{Daniel T. Speckhard,\textit{$^{a, b}$} Tim Bechtel,\textit{$^{a, b}$} Luca M. Ghiringhelli,\textit{$^{c}$} Martin Kuban,\textit{$^{a}$} Santiago Rigamonti,\textit{$^{a}$} and Claudia Draxl\textit{$^{a, b}$}} \vspace{0.3cm} \\

& \noindent\normalsize
{
}
\end{tabular}
 \end{@twocolumnfalse} \vspace{0.6cm}
  ]

\renewcommand*\rmdefault{bch}\normalfont\upshape
\rmfamily
\section*{}
\vspace{-1cm}

\footnotetext{\textit{$^{a}$~Physics Department and CSMB, Humboldt-Universit\"at zu Berlin, Zum Gro\ss en Windkanal 2, 12489 Berlin, Germany, Fax: +49 2093 66361; Tel: +49 2093 66363; E-mail: claudia.draxl@physik.hu-berlin.de}}
\footnotetext{\textit{$^{b}$~Max Planck Institute for Solid State Research, Heisenbergstraße 1, 70569 Stuttgart, Germany}}
\footnotetext{\textit{$^{c}$~Department of Materials Science and Friedrich-Alexander Universität Erlangen-Nürnberg, Dr.-Mack-Str. 77, 90762 Fürth, Germany}}

\begin{abstract}
Big data has ushered in a new wave of predictive power using machine learning models. In this work, we assess what {\it big} means in the context of typical materials-science machine-learning problems. This concerns not only data volume, but also data quality and veracity as much as infrastructure issues. With selected examples, we ask (i) how models generalize to similar datasets, (ii) how high-quality datasets can be gathered from heterogenous sources, (iii) how the feature set and complexity of a model can affect expressivity, and (iv) what infrastructure requirements are needed to create larger datasets and train models on them. In sum, we find that big data present unique challenges along very different aspects that should serve to motivate further work.
\end{abstract}

\section{Introduction}
\textit{Big Data} is a term that not only governs social media and online stores, but has entered most modern research fields. As such, it also concerns materials science. While our research has always been largely based on data, through the analysis and interpretation of measured and computed results, in recent times, more aspects have come into play. For instance, there is the practical reason that more and more funding bodies require data to be kept for a certain period of time. On the scientific side, the idea of sharing data is enjoying popularity, thus avoiding that the same investigation is done multiple times. Most important is the use of data in machine learning (ML), or more generally, artificial intelligence (AI).

Artificial Intelligence is arguably the most rapidly emerging topic in various research domains, with increasing impact in materials science as well. The success of AI approaches, however, strongly depends on the amount and quality of the underlying training data. In this context, the first obvious question is how large a dataset needs to be in order to provide sufficient information for the research problem to be solved. Are millions of scientific results, as available in international databases, a gold mine? Or do these collections still not provide enough significant data for a specific question? Or can we even learn from small datasets? Are large, multipurpose databases sufficient for training models, or to what extent are dedicated datasets needed for this task? Can errors be controlled when using data from different sources? Can we learn from experimental and theoretical data together, or are even different theoretical or experimental data by themselves too heterogeneous to produce reasonable predictions? 

What is useful for a particular learning task, may depend heavily on the research question, the quantity to be learned, and the methodology employed. This concerns data quality, data interoperability -- especially when data from different sources are brought together -- data veracity, and data volume (the last two are part of the "Four V's of Big Data"~\cite{draxl2020big}). In this work, we ask the question: "What does {\it Big} mean in the realm of materials science data?". There are many issues related to this, demonstrated by a few examples: First, some methods are more data hungry than others. While, for instance, symbolic regressors may lead to reasonable results already for a small data set \cite{speckhard2023extrapolation}, neural networks (NNs) may require many more data points~\cite{jha2018elemnet} for training to be stable. Second, even with what can be considered big datasets, training models that generalize to similar datasets is not trivial. Third, data quality may have a significant impact. While high-quality data may give one a clear picture from the very beginning, many more {\it noisy} data may need to be accumulated for obtaining robust results, to avoid wrong conclusions and/or allow for physical interpretation. Finally, what does Big Data mean for data infrastructure? What are the requirements for processing and storing big datasets; how compute-intensive is training ML models on them?

In this work, we address such questions and evaluate the performance of different AI models and tools in terms of data volume, diversity, and quality and provide a quantitative analysis. As the overall topic is very wide and diverse, we aim to draw the reader's attention to it and initiate discussions rather than to provide detailed solutions to all of the related issues. The chosen examples should serve this purpose. In the first one, we ask whether a model obtained from a message-passing neural network (MPNN) can be transferred to a similar dataset. Second, similarity metrics and sorting are applied to understand data quality in terms of convergence behavior of density-functional theory (DFT) data. Third, the effect of an expanded feature set on the expressivity of cluster expansion is demonstrated. Fourth, the complexity of different model classes is explored in terms of their performance. Finally, infrastructure requirements for creating large materials datasets and for training models with many parameters are quantitatively analyzed.

\begin{figure*}[h!]
\centering
\includegraphics[width=0.6\textwidth]{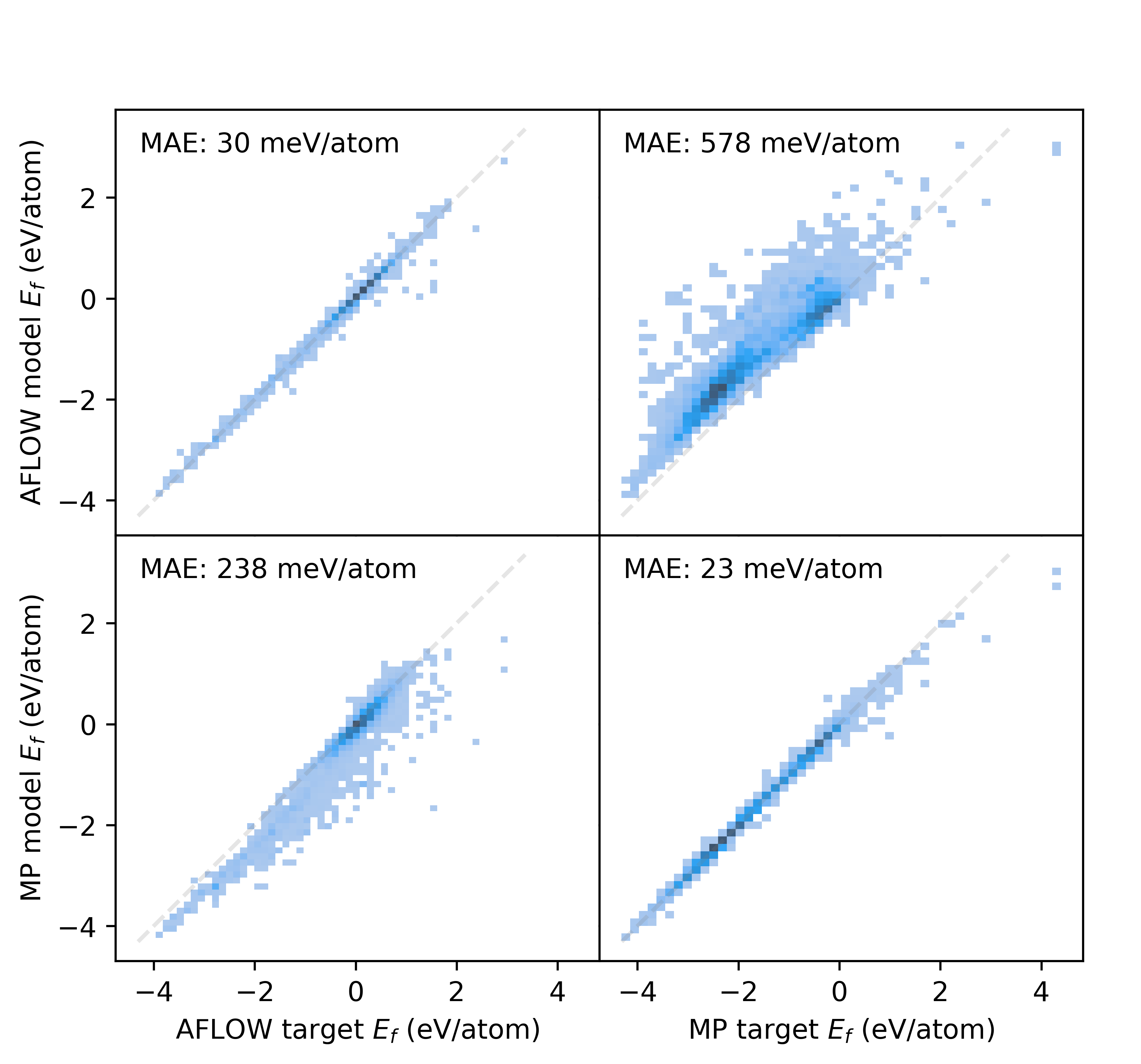}
\caption{Predicted versus calculated formation energies for AFLOW and Materials Project (MP) data. Top-left: model trained and tested on AFLOW data; top-right: model trained on AFLOW and tested on MP; bottom-left: model trained on MP and tested on AFLOW; bottom-right: model trained and tested on MP.}
\label{fig:combined regression}
\end{figure*}

\section{Transferability of models}
\label{sec:transferability_models}
Materials Project (MP) \cite{MP} and AFLOW \cite{AFLOW} are popular data collections from computational materials science. Both are also often used for ML purposes. Most of the calculations are carried out by DFT with the same code (VASP, projector augmented wave method) and exchange-correlation functionals (mostly PBE). Since both databases also contain a significant amount of materials which are common to both, \ie share the same composition and spacegroup, we ask the question whether a model trained on one dataset would perform well when making inferences on the other dataset. We choose the formation energy as our target, since it is a rather well-behaved property, only relying on the total energies of the compounds and their constituents. 

The AFLOW dataset is chosen as described in Ref.~\citenum{bechtel2023band} and filtered to use only calculations performed with the PBE functional, which contain both bandgap and formation energy. The MP dataset is retrieved from Ref.~\citenum{chen2019graph} (dataset snapshot denoted as \textit{MP-crystals-2018.6.1}). After filtering, the two datasets comprise 61,898 and 60,289 materials, respectively. We proceed by assigning unique identifiers to the structures in the AFLOW and MP datasets. They consist of the chemical formula concatenated with the space group of the crystal, similar to Ref.~\citenum{schmidt2021crystal}, to define which structures are common to both databases and which are unique to either of them. This results in 50,652 unique composition-spacegroup pairs in the AFLOW dataset and 54,438 in the MP dataset, of which 8,591 pairs are common to both datasets.

To avoid leakage of information from the training set to the test set, we ensure that the same common structures are present in the training-test splits of both datasets. This means that a model trained on AFLOW data will not be used for inference on MP test data on the same structure it was trained on, and vice versa. For instance, all structures with composition Mg$_2$F$_4$ and spacegroup 136 get the label Mg2F4\_136. If there are several of these structures, \eg with different lattice parameters, all of them get assigned to one split, \ie training, validation, or test.

\begin{figure}[h]
\centering
\includegraphics[scale=0.5]{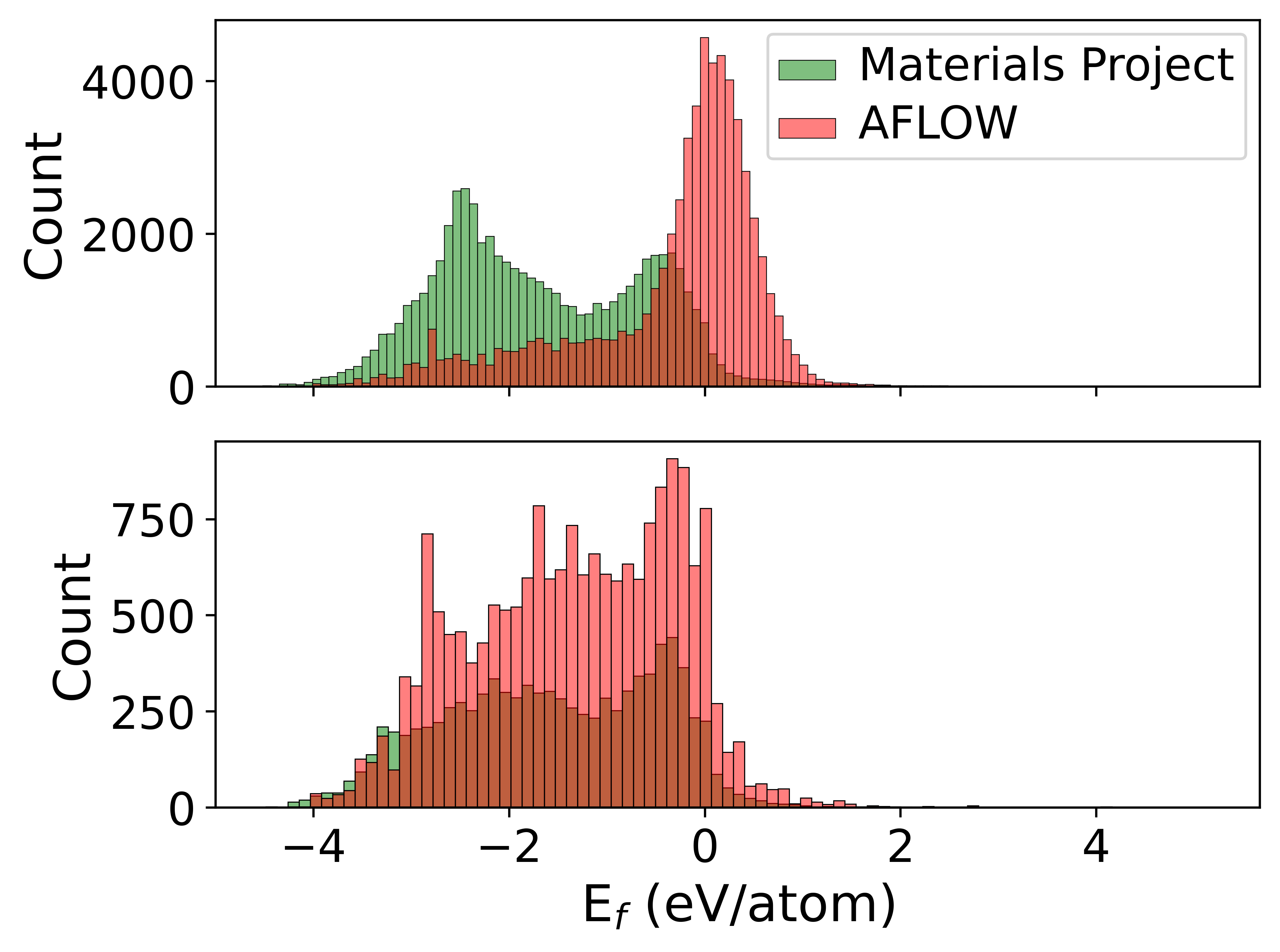}
\caption{Top: Distribution of formation energies in the two datasets containing 50,652 (AFLOW) and 54,438 (MP) unique composition-spacegroup pairs. Bottom: Distribution of formation energies for the 8,591 composition-spacegroup pairs that are present in both datasets.}
\label{fig:hist}
\end{figure}

We train MPNNs with edge updates, as described in Ref.~\citenum{jorgensen2018neural} on the two datasets. Previously, we have shown that the hyperparameters that define the architecture of the MPNN transfer well to different prediction tasks~\cite{bechtel2023band}. Following this, we perform no further hyperparameter optimization.

Figure \ref{fig:combined regression} shows the performance of the two models on the two datasets. While the predictions are very good when training and test data are from the same database -- mean absolute errors (MAE) of 30 meV and 23 meV for AFLOW and MP, respectively, the errors in the prediction of the other database are an order of magnitude larger. Notably, there is also a clear trend of too large (small) predictions on the upper right (lower left) of the matrix. This points to a systematic offset between the predicted energies of the two datasets. We can partly trace back the discrepancies to the relatively small overlap of the spacegroup-composition pairs in the two databases, which is about 16\%. Further analysis, indeed, reveals that MP data contain many more materials with lower formation energies than the AFLOW data as evident from the top panel of Fig. \ref{fig:hist}. Differences may arise, however, also from computational details, such as Brillouin-zone (BZ) sampling, basis-set cutoff, convergence criteria, etc. The distribution of the shared structures (bottom panel of Fig. \ref{fig:hist}) indicates that the latter have overall a smaller impact, \ie do not lead to a systematic offset. However, we see that there are many more entries of the same structures, \eg with different lattice parameters, in the AFLOW dataset. For example, Ti$_2$O$_4$ (spacegroup 136) and BaTiO$_3$ (spacegroup 221) occur 127 and 115 times, respectively, while only one of each compound is found in the MP data.

Even for what we might consider to be big datasets in materials science, and using a rather simple ML target, this example shows us that the models we train are only valid for the data they were trained on and struggle to generalize. In our specific example, the AFLOW and MP datasets turn out to sample the underlying material space differently, since the MP materials appear biased towards lower formation energies. We conclude that these two databases are not "big" enough in the sense that they are not diverse enough to be able to make predictions across a wider set of diversity. Training on the combined dataset, the predictions are less biased \ie the errors being more symmetric with respect to the diagonal. However, the MAE is higher overall as compared to the individual models. This indicates that also differences in computational settings may play a role and may need to be considered as input features.

\section{Revealing data quality by similarity measures}
\label{sec:similarity}
Veracity, another of the "Four V's of Big Data" poses a challenge to ML applications by introducing a fundamental level of noise that cannot be overcome with more complex models. An example from computational materials science is sets of calculations with different levels of accuracy\cite{lejaeghere2016reproducibility} determined by the chosen approximation, such as the exchange-correlation (xc) functional of DFT, or different levels of numerical precision, determined by input parameters, such as basis-set size or $k$ grid. Related to this, is a practical problem: If multiple calculations from different sources --with smaller or larger deviations in the results for a particular physical property-- are available, which value should be trusted? This challenge can be met by either applying corrections to the data to bring them onto the same level of accuracy/precision~\cite{huang2020quantum} or by filtering them for consistent subsets. The former option requires additional ML models \cite{speckhard2023extrapolation}, trained on dedicated, high-precision datasets. However, generating the required data for training these models is costly, and making predictions for materials with large unit cells may require that the models are trained on such systems as well. Filtering the data by their numerical precision, on the other hand, can be applied to existing datasets, but one may not find enough data for a particular application because the number of calculations with exactly the same numerical settings is typically small. The number of calculations that can be used together can be increased, however, if we can understand --and quantify-- the level of convergence of data with the computational parameters. Understanding and visualizing the convergence behavior of computed properties can be achieved by introducing similarity metrics, like recently demonstrated with the example of fingerprints of the electronic density of states (DOS) \cite{Kuban2022MRS, Kuban2024MADAS}. Given a large enough set of calculations for a single material, the relationships between the parameters used in calculations and the similarities of the results can be shown.

\begin{figure*}[!h]
\centering
\includegraphics[width=0.66\textwidth]{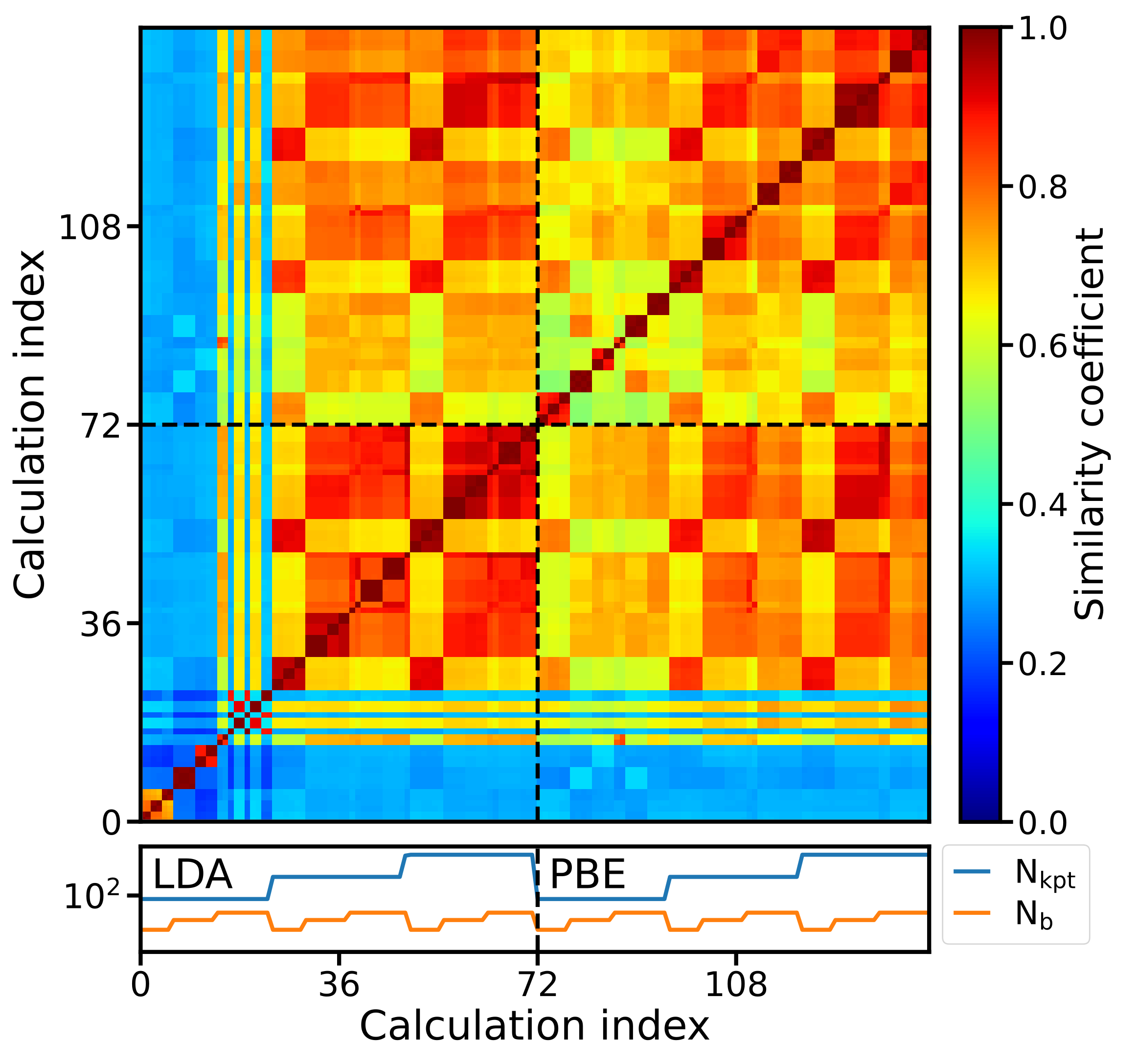}
\caption{DOS similarity matrices for h-BN obtained with different basis-set sizes and $k$-meshes, and two different exchange-correlation (xc) functionals. The data are sorted by the latter, where low indices ($\leq$ 71) correspond to the local-density approximation (LDA), high indices (>71) to the generalized-gradient functional PBE. The color code indicates the similarity coefficient. The bottom panel shows the number of $k$-points (blue) and the number of basis functions (orange).}
\label{fig:DOS_similarity_XC_effects}
\end{figure*}

To illustrate this, we make use of data from the NOMAD data collection~\cite{NOMAD,CarbognoFHIaimsErrorsData}. These calculations were carried out with the DFT code FHI-aims~\cite{Blum2009FHIaims} as part of a systematic study of the impact of computational parameters on DFT results~\cite{carbogno2022numerical}. For our analysis, we use the calculations of hexagonal boron nitride (h-BN), specifically, the DOS of ground-state calculations obtained with different basis-set sizes and $k$-point samplings at the experimentally determined equilibrium volume. To compare the results of these calculations, we compute the spectral fingerprints~\cite{Kuban2022SDATA} of the DOS in the energy range between $-10$ and $10$ eV around the Fermi energy and calculate the corresponding similarity matrix. The rows and columns are sorted by increasing numerical settings, separately for the two xc functionals LDA (matrix indices $\leq 71$) and PBE (indices $> 71$). This matrix is depicted in Fig. \ref{fig:DOS_similarity_XC_effects}. The bottom panel shows the number of $k$-points (N$_\mathrm{kpt}$) and basis functions (N$_\mathrm{b}$). For each $k$-point mesh (plateau in the $k$-mesh), the basis set is increased in the same way. Additionally, the data are sorted by a set of numerical settings, called "light", "tight", and "really tight" in FHI-aims. Finally, consecutive calculations with otherwise identical settings vary by the relativistic approximation employed for core electrons, \ie "ZORA", and "atomic ZORA"\cite{Blum2009FHIaims}. Sorting the matrix in this way, we see patterns appearing in the figure, which we discuss below.

Focusing first on the convergence of the LDA data (indices $i\leq71$), we observe a clear block structure, where the calculations of the first set, \ie those with the lowest N$_\mathrm{kpt}$ (index $i\leq23$) are most dissimilar to all others, and also to each other. However, they are pairwise similar, indicating that the relativistic approximation plays a minor role in the convergence of the DOS. Sets of calculations with medium ($24\leq i \leq 47$) and high ($48\leq i\leq 71$) numbers of $k$-points generally show higher similarity among themselves. Noticeably, calculations with a low (medium and high) number of basis functions N$_\mathrm{b}$ are similar among different BZ samplings, but less similar to calculations with higher (lower) N$_\mathrm{b}$. For PBE calculations, the convergence behavior is different: Already calculations with low N$_\mathrm{kpt}$ reach medium similarity to calculations with high N$_\mathrm{kpt}$. Contrary to the LDA data, PBE calculations with low and medium N$_\mathrm{kpt}$ don't reach high similarity to PBE calculations with highest settings, indicating that PBE calculations require a higher $k$-point density for this material. Further comparing the similarity of LDA and PBE convergence, \ie the off-diagonal blocks of the matrix (indices $i\geq71$ on the x-axis and $i\leq72$ on the y-axis), we find a pattern corresponding to calculations with the same number of basis functions. It shows that calculations with different xc functionals behave more similar to each other if the same number of basis functions are used.

Our example illustrates the veracity of materials data arising from the large combinatorial space of approximations and numerical parameters employed in DFT codes. While LDA and PBE, both being semi-local functionals, are generally expected to give similar results in terms of the electronic structure, we show here that the convergence behavior with the number of $k$-points and basis functions, is surprisingly different. The visualization of DOS similarity matrices allows one to understand and quantify differences between DFT data computed with different computational settings. It also enables the creation of rules to group data in order to gather "homogeneous" subgroups within a dataset.

\section{Striving for expressitivity in cluster expansion}

\begin{figure*}[h!]
\centering
\includegraphics[scale=0.7]{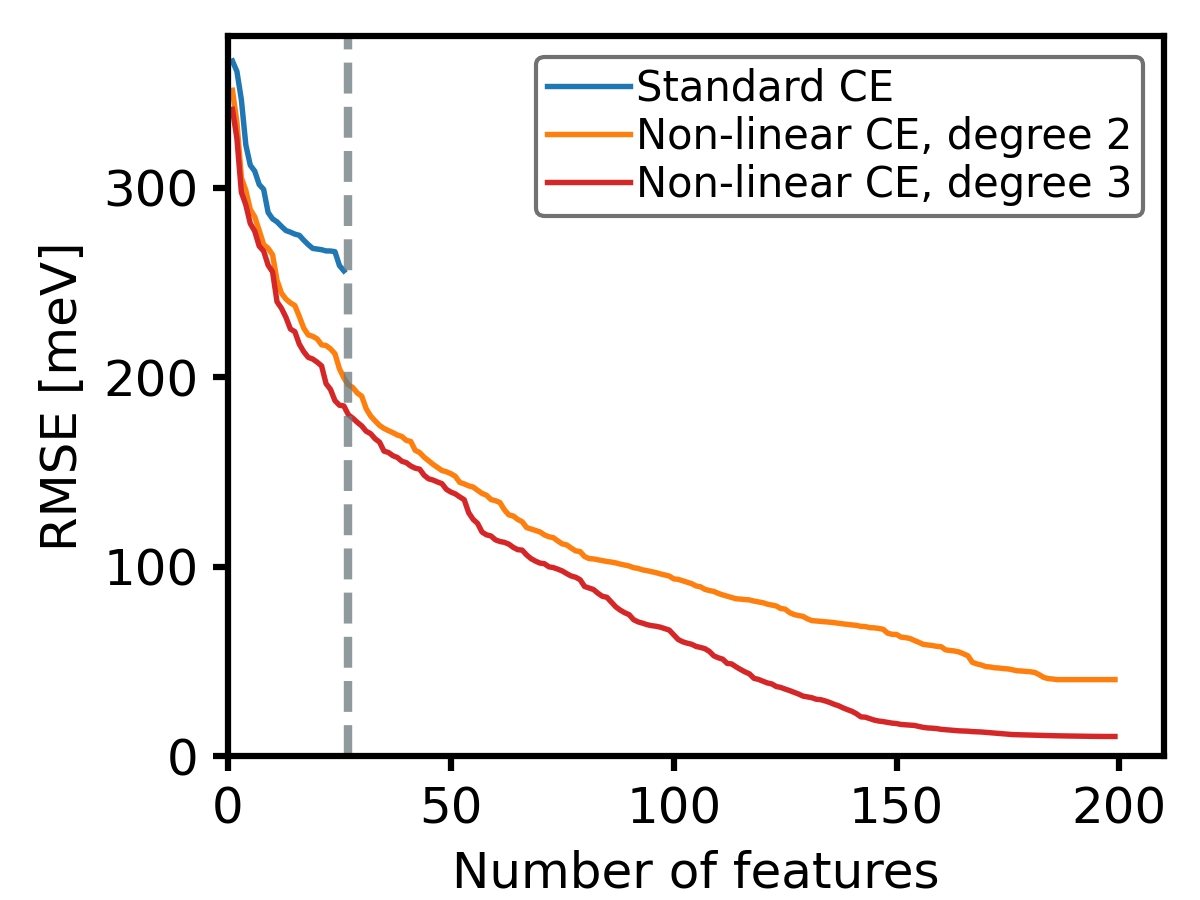}
\includegraphics[scale=0.7]{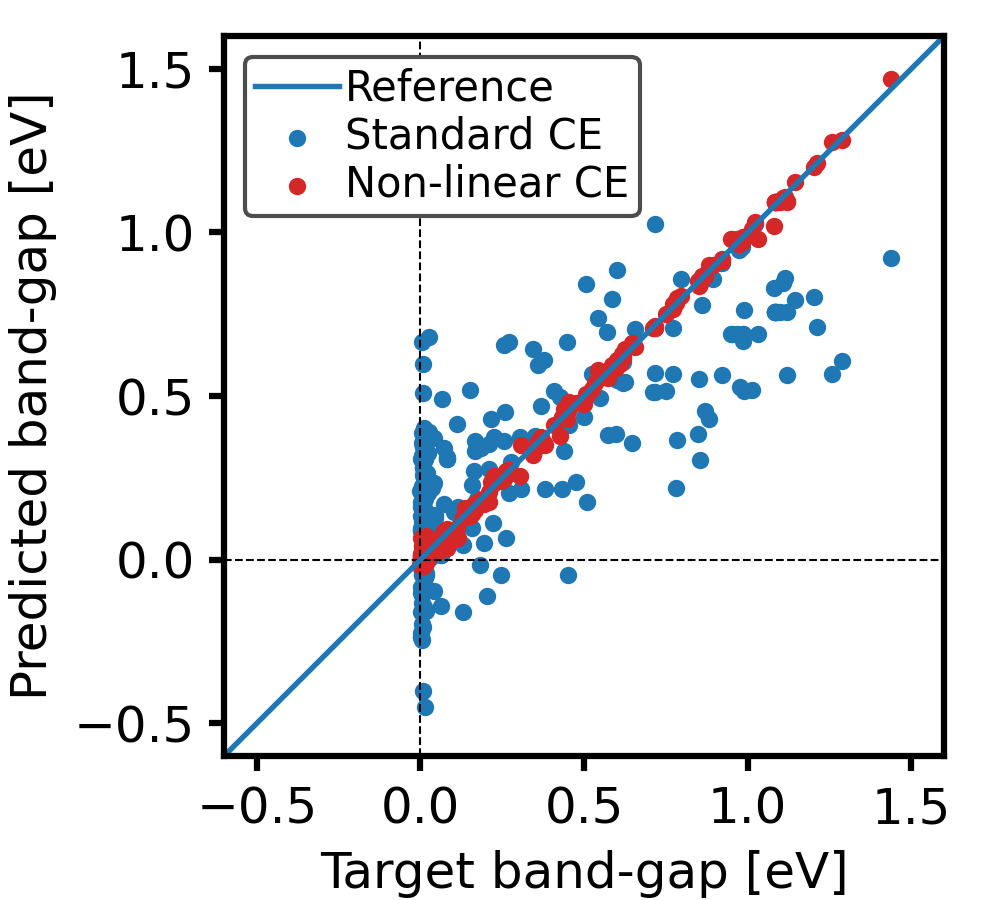}
\includegraphics[scale=0.7]{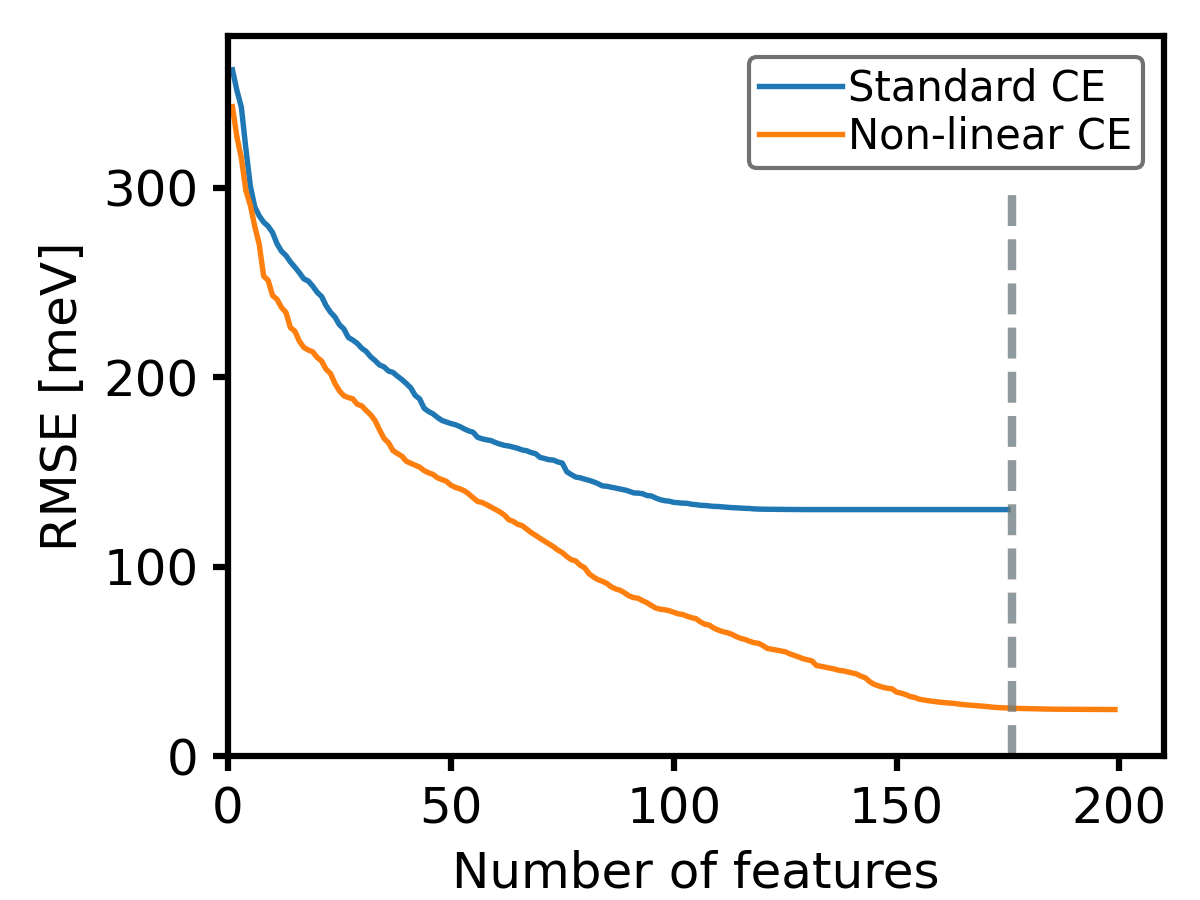}
\includegraphics[scale=0.7]{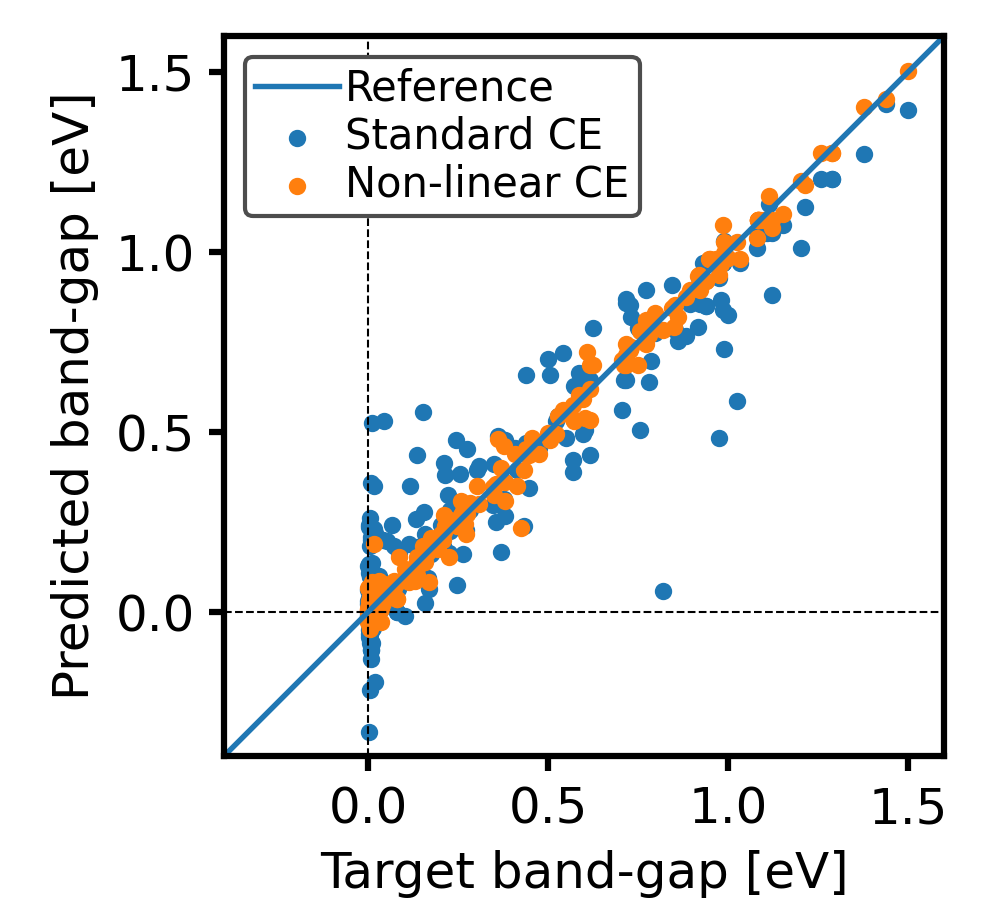}

\caption{Left: Fitting errors for standard (blue) and nonlinear CE models (orange, red) based on a pool of 27 clusters (top) and 176 clusters (bottom). Right: Predicted versus target bandgaps for the standard CE with 27 clusters and the nonlinear CE of degree 3 models with 150 features (top); and standard CE and nonlinear CE models with 150 clusters / features (bottom). }
\label{fig:nlce}
\end{figure*}

In this example, we investigate the role of the feature set in the modeling of the bandgap of a family of perovskite materials by means of the cluster-expansion (CE) technique. CE is a multi-scale method based on a lattice Hamiltonian~\cite{Sanchez1984}, which is used to model and predict properties of alloys at different compositions and temperatures. In CE, an alloy configuration is represented by a vector $\bm{s}$ of occupation variables $s_i$. For a binary alloy, for example, the occupation variables may take the values $s_i = -1$ or 1, indicating the presence of one or the other species at crystal site $i$. By defining the cluster functions $\Gamma_c(\bm{s})=\prod_{i\in c}s_i$, where $c$ is a subset of crystal sites or {\it cluster}, it can be demonstrated \cite{Sanchez1984} that any configuration-dependent property, can be formally expanded in terms of the symmetry-averaged cluster functions $X_c(\bm{s})=\langle\Gamma_c(\bm{s})\rangle$, named {\it cluster correlations}, where $\langle\cdot\rangle$ represents the symmetry average over all clusters symmetrically equivalent to $c$. For modeling the bandgap, this reads
\begin{equation}
    E_g(\bm{s}) = \sum_c \lambda_c X_c(\bm{s}).
\end{equation}
Here, the sum runs over a set containing $p$ symmetrically inequivalent clusters. In what follows, we call this family of models "Standard CE". Cluster expansion can be viewed as a ML problem \cite{rigamonti2023cell,stroth2024nonlinearce}, where the cluster correlations $X_c$ are input features describing the structure of the crystal. This view can be leveraged by considering basis expansions based on these features, leading to nonlinear CE models \cite{stroth2024nonlinearce} defined as  
\begin{equation}
    E_g(s) = \sum_m \theta_m h_m(\bm{X}(s)),
\end{equation}
with the (possibly nonlinear) functions $h_m(\bm{X}):\mathbb{R}^p\rightarrow\mathbb{R}$ and $\bm{X}$ a $p$-vector with components $X_c$. The sum runs over a set containing $q$ functions $h_m$. For this example, we choose nonlinear polynomial features: Every function $h_m(\bm{X})$ is a monomial up to degree $d$, \eg $h_j(\bm{X})=X_1^2X_2$. The total number of features in this case is given by $\sum_{k=0}^d\binom{p + k - 1}{k}$~\cite{stroth2024nonlinearce}.  

In this section, we want to explore the expressiveness of the feature spaces in standard and nonlinear CE. Assuming that the intrinsic error in the data to be modeled is small, does our feature space allow us to approximate the ground truth? We use a dataset of 235 oxide perovskites with composition BaSnO$_{3-x}$, with $x<1$ being the number of oxygen vacancies per formula unit. BaSnO$_3$ has significant potential for use in photocatalytic and optoelectronic applications. Its electronic structure demonstrates a complex dependence on the concentration and configuration of oxygen vacancies. The bandgap varies strongly even among structures with identical concentration and comparable energies of formation. Thus, modeling the bandgap is challenging for linear regression models. For this learning task, the clusters account for specific structural patterns of O vacancies.

As shown in Fig. \ref{fig:nlce}, standard CE models obtained by orthogonal matching pursuit (OMP) from a pool of $p=27$ clusters, yield fit root-mean-squared errors (RMSEs) larger than 250 meV (blue line, top-left panel). The best fit is obtained, as expected, when using all 27 features (marked by a vertical dashed line). For the pool of $p=27$ clusters, we explore polynomials of degree $d=2$ and $d=3$, containing 405 and 4059 features, respectively, obtaining models with up to 200 features using OMP. These are shown by the orange and red lines in the top-left panel of Fig.\ref{fig:nlce}. In addition to the expected monotonous reduction of the error, two notable findings are obtained: first, with 27 features, the nonlinear feature space yields better models than the standard CE with the same number of 27 clusters. This is remarkable because the nonlinear features of the nonlinear CE models are derived solely from the 27 cluster correlations used in the standard CE model. Related to this point, we also observe that for the same number of features, degree 3 always gives smaller error than degree 2. Second, both nonlinear CE models reach a point where no further improvement is obtained upon increasing the number of features. This is observed as a plateau above $\sim 150$ features for $d=3$ and above $\sim 180$ features for $d=2$. The plateau for $d=2$ is reached at a higher level of the RMSE than for $d=3$. Thus, under the assumption that the intrinsic error is small, increasing the complexity of the feature space allows the CE to better approach the underlying ground truth. In practice, increasing the complexity of the feature space allows CE to obtain the same accuracy with sparser models as is obtainable with less complex feature spaces. This is important for model selection methods that favor sparsity, such as LASSO. The top-right panel of Fig. \ref{fig:nlce} shows the quality of the predictions for the whole dataset using the best possible model for standard CE (blue dots) and the best possible model for $d=3$ nonlinear CE, based on the same 27 cluster correlations. The improvement of the nonlinear modeling is remarkable, especially for the difficult case of metals (the set of materials with $E_g=0$).

Now, the question arises whether the beneficial effect of nonlinear features could be also obtained by adding more cluster correlations, that is, accounting for more structural patterns of vacancies. For this, we have created a pool of 176 clusters. The result of standard CE is shown in the lower left panel of Fig.~\ref{fig:nlce} (blue line). Like before, we see a plateau setting in at about 110 clusters with a RMSE of $\sim 125$ meV, and no further improvement can be achieved upon adding more clusters. The vertical dashed line indicates the model with the largest possible number of clusters in this pool. Again, the second-order nonlinear CE based on this same pool allows for obtaining better fits at all numbers of features. A comparison of the quality of the fits for standard and nonlinear CE models with 150 clusters, respectively, is shown in the lower-right panel of the figure. Similarly to what was obtained for the small pool of 27 clusters, the nonlinear CE yields significantly better predictions.

From this example, we see the feature set can have a strong impact on the expressiveness of the CE model. Including nonlinear terms enables the CE to better fit the underlying ground truth even with a small data set size. This benefit appears irrespective of the number of cluster correlations. This improvement motivates us to to try to better define model complexity which is done in the following section.

\section{Model complexity}

Can knowing the complexity of the model tell us something about model performance for a given dataset? The term model complexity is not well defined in the literature. It  describes the capacity of a model to learn an underlying probability distribution. We start this section by defining what complexity means for several model classes. We then evaluate how complexity determines model performance for two example datasets. Finally, we discuss the search for a model-independent scalar that represents model complexity.

\label{sec:data_volume}
\subsection{Complexity for neural networks, random forest, and SISSO models}
Here, we discuss how complexity can be quantified for three different types of models, \ie neural networks (NN), the sure independence screening sparsifying operator (SSISO)~\cite{ouyang2018sisso}, and random forests (RF). For NNs, the total number of trainable parameters is often used as a measure for the model's complexity. That said, some of the NN parameters have very different impact. The output of node $j$ at layer $k+1$ in a feed-forward network is:
\begin{equation}
x^{k+1}_j = \sum_i \sigma(W_{ij} x^k_i + b_j).
\label{eq:neural_network}
\end{equation}
This node receives an input $x$ from layer $k$, which is multiplied by a weight parameter $W_{ij}$ and summed with a bias parameter $b_j$ before being fed into a nonlinear activation function, $\sigma$. Thereby the bias parameter is quite different from the weight parameter. So, a better description of complexity for NNs is a two-dimensional vector, containing the number of weights and biases. The benefit of this definition is that it is simple to compute. The drawback is that it is not a unique definition and does not differentiate between parameters deeper in the network, which are more expressive because they transform nonlinear inputs with nonlinear functions~\cite{nn_complexity}.

The SISSO model first combines {\it primary} features into {\it generated} features using a set of mathematical operations (\eg $+, -, \exp(), \sqrt, \ldots$, and combinations thereof) \cite{ouyang2018sisso,purcell2023recent}. The number of features that are selected by the algorithm is called the dimension of the model. We can interpret this in terms of a (symbolic) tree that describes a generated feature, where each split in the tree is an operation and the leaves are the primary features \cite{purcell2023recent}. The larger the depth of the tree, the more operations there are in the generated-features space. The maximum tree depth is called the \textit{rung} of the model. In essence, the number of selected generated features (model dimension) defines how many coefficients the model has. One can also include a single constant bias term in SISSO. With a larger rung value, the possible secondary features explode combinatorially. Therefore, we can define a descriptor for the complexity in SISSO with a two-dimensional vector, comprising the rung and the dimension of the model (counting the bias term if present as an extra dimension).

How can we define complexity for RFs? RFs are piecewise constant functions. For a regressor, each leaf node in each decision tree learns a constant bias term. For each split in each decision tree, a value of the variable to be split on must be learned. Since the trees are binary, the number of leaf nodes is one greater than the number of splits. Therefore, the number of leaf nodes is an integer that tells us the number of trainable parameters in the model and describes the complexity of the model.

From each of these three model types, we have presented a measure of model complexity. The model complexity for each model type is described with a different number/vector. For NNs, it is a two-dimensional vector of the total number of weights and biases; for SISSO, it is a two-dimensional vector reporting rung and dimension of the model, while for RFs, it is the number of leafs in the model. The definitions are not unique but offer an idea of how well the model can learn to approximate a wide variety of functions. 

\subsection{Performance as a function of complexity}

Can model performance be predicted by the complexity of the model? Two examples are examined here using the models from the previous section. In the first example, an NN and a three-dimensional SISSO model are fit to the DFT data used in Ref.~\citenum{carbogno2022numerical} using 80\% of the data for training and 20\%  for testing. The SISSO model outperforms the best performing NN architecture with a test root-mean-squared-logarithmic error of 0.231 compared to 1.367, despite having less parameters. The NN complexity can be scaled by changing the number of nodes and layers in the model. An NN with the same number of linear, nonlinear, and constant parameters as the SISSO model still does an order of magnitude worse than the SISSO model. This may not comes as a surprise, since NNs are considered to be data hungry~\cite{aggarwal2018neural}. From this example, however, we can deduce that the number of constant, linear, and nonlinear parameters in the model is not enough information to predict model performance. Rather, the model class is the determining factor here.

In the second example, an NN and an RF are trained to predict bandgaps on the AFLOW dataset of Ref.~\citenum{bechtel2023band}. Features describing the elemental composition of the materials, as described in Ref.~\citenum{ward2016general}, are fed into both models. For the same parameter count, one can argue that an NN with the ReLU activation function is more complex than the RF, since the RF model is piecewise constant and the NN is piecewise linear. However, both NNs and RFs are universal approximators, meaning that with enough splits/nodes, they can approximate any probability distribution. On the AFLOW bandagap dataset, after cross-validation, the RF outperforms the NN, with a test MAE of 469 meV and 515 meV respectively. This trend holds when we restrict the total number of parameters of both models to be similar. Once again, model class, not model complexity, is the decisive factor of model performance.

We can conclude from both examples that the number of trainable parameters alone tells us very little about model performance. Rather, we find that for a given dataset, the performance depends on the model class.

\subsection{Searching for a definition of model complexity}

Can model complexity be defined with a scalar? In the previous subsections we saw that the total parameter count is a poor metric for NNs. A similar argument can be made for polynomial regressors, where some parameters allow for linear terms to be used by the model and others allow for nonlinear terms. One could therefore be motivated to count the coefficients to linear and nonlinear terms and place them on uneven footing, since the nonlinear coefficients give the model a different type of flexibility than the linear coefficients. The question though is, how to weight these terms to come up with a more general metric for model complexity. The simplest combination would be a weighted sum:
\begin{equation}
\text{C(h)} = a*\text{A(h)}+ b*\text{L(h)}+c*\text{N(h)},
\label{eq:complexity}
\end{equation}
where C is the complexity of the model h, A is the number of additive parameters, L is the number of multiplicative coefficients to linear terms in the features, and N is the number of coefficients of nonlinear combinations. If we set the coefficients of the complexity metric, $a$, $b$, and $c$ to unity, we recover the total trainable parameter count. Assigning values to $a$, $b$, and $c$ is however not an easy task. The proposed complexity metric will mean different things for different models. As discussed, RFs have additive constant parameters that need to be learned, but no coefficients to linear or nonlinear combinations of the features. So with 10,000 additive constant parameters the RF can learn to approximate many different distributions with a low mean-squared error. A generalized linear model, however, is only able to approximate constant functions with constant additive parameters. Therefore, with the same value of complexity, as defined in Equation \ref{eq:complexity}, the RF has a much higher capacity to learn a wide variety of distributions than the generalized linear model.

These examples demonstrate that the total parameter count is not sufficiently descriptive of the model's ability to learn to approximate a wide variety of functions for many model classes. We offer an alternative metric for model complexity as a weighted sum of constant, linear, and nonlinear terms. We conclude however, that the meaning of complexity is model dependent. This motivates us to therefore only talk about complexity within a single model class. More work is needed to explore these topics more carefully from an information theoretic point of view.

\section{Infrastructure requirements}

\subsection{File and storage requirements for dataset calculation}
\begin{figure}[h]
\centering
\includegraphics[scale=0.6]{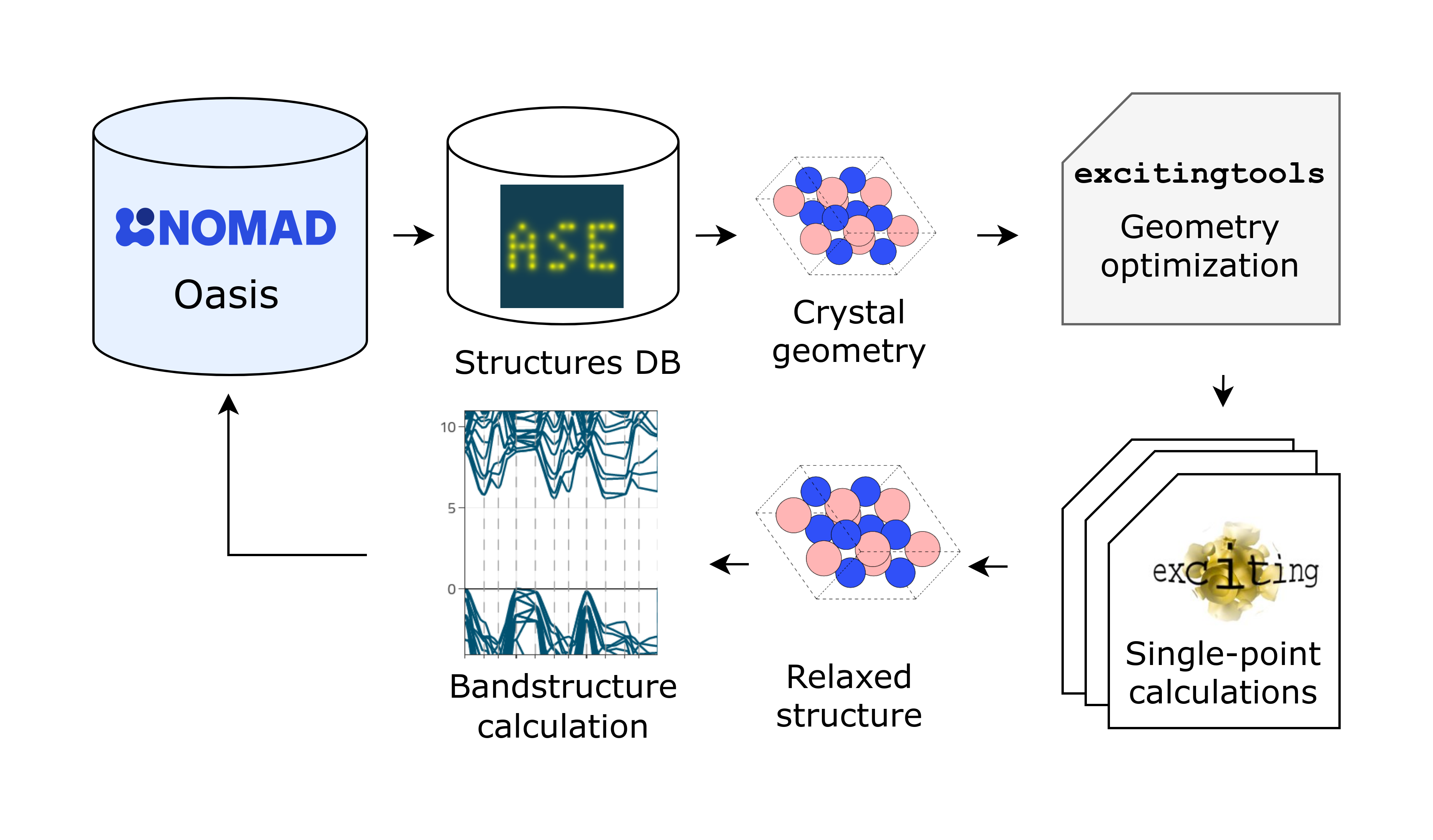}
\caption{Workflow for high-throughput geometry optimization, followed by a bandstructure calculation. Crystal structures are pulled from a NOMAD Oasis instance and stored in an ASE database (DB). The structures are read by the Python API \excitingtools, and a geometry optimization is carried out, consisting of multiple single-point ground-state calculations with \exciting. For the resulting relaxed geometry, a bandstructure calculation is performed. All output files are uploaded to a NOMAD Oasis instance.}
\label{fig:workflow}
\end{figure}

In order to create large curated datasets of DFT calculations, sophisticated high-throughput workflows have become an indispensable tool. Here, we would like to provide an estimate for what this means in terms of infrastructure requirements. In Fig. \ref{fig:workflow}, an example of such a workflow is depicted, leaving out workflow managers~\cite{rosen2024jobflow}, validity constraints~\cite{schintke2024validity}, and alike for the sake of simplicity. Here, an Atomic Simulation Environment (ASE)~\cite{larsen2017atomic} database is employed to store input crystal structures that the user desires to simulate with \exciting, an all-electron code based on the linearized augmented planewave method~\cite{gulans2014exciting}. In this specific workflow, each structure is relaxed, \ie the geometry is optimized by running multiple single-point calculations, via the open-source Python API \excitingtools~\cite{buccheri2023excitingtools}. The bandstructure is calculated afterwards for the equilibrium geometry. This rather simple workflow creates 41 output files and roughly 30 MB of data for each crystal structure. In addition, for each structure, the calculation must be converged with respect to BZ sampling and basis-set size. Typical usage would see at least three different $k$-point grids and three different basis-set sizes. For 30,000 crystal structures, which is not too big a number for ML tasks, (see, \eg section \ref{sec:transferability_models}), and nine different settings per structure, we perform 270,000 relaxations and bandstructure calculations. This amounts to 11,070,000 files, or in terms of storage, roughly 8.1 TB of data. 

Such workflows are typically executed on supercomputers, which impose limits on users concerning disk space and numbers of files. Storage limits are, \eg on the order of a few 100 GB in backed-up directories, or a few TB in scratch directories. There are also limits in terms of files that can be stored (often at the most 1 million). Both limits often mean that the users need to compress their data periodically and transfer their files to a storage system outside of the supercomputer network. In this example, the data are transferred to a NOMAD Oasis~\cite{nomad_oasis_webpage,scheffler2022fair}. NOMAD Oasis is the same software as in the public NOMAD data infrastructure \cite{Scheidgen2023}, including among other tools, parsing, normalizing, electronic notebooks, and the NOMAD AI toolkit~\cite{sbailo2022nomad}. The software can be installed locally by any research group and can be configured to their needs. Calculated data, as described above, are stored into an interoparable format~\cite{draxl2019nomad,ghiringhelli2023shared}.

In summary, calculating big data presents its own unique infrastructure challenges. The workflow example shown here quantitatively demonstrates that performing high-throughput DFT calculations requires sophisticated hardware/software solutions to navigate HPC file and storage restrictions. The NOMAD Oasis is one such solution to this problem, which also allows for making data ready for ML purposes.

\subsection{Computing requirements for training large models}

\begin{figure}[h]
\centering
\includegraphics[scale=1.]{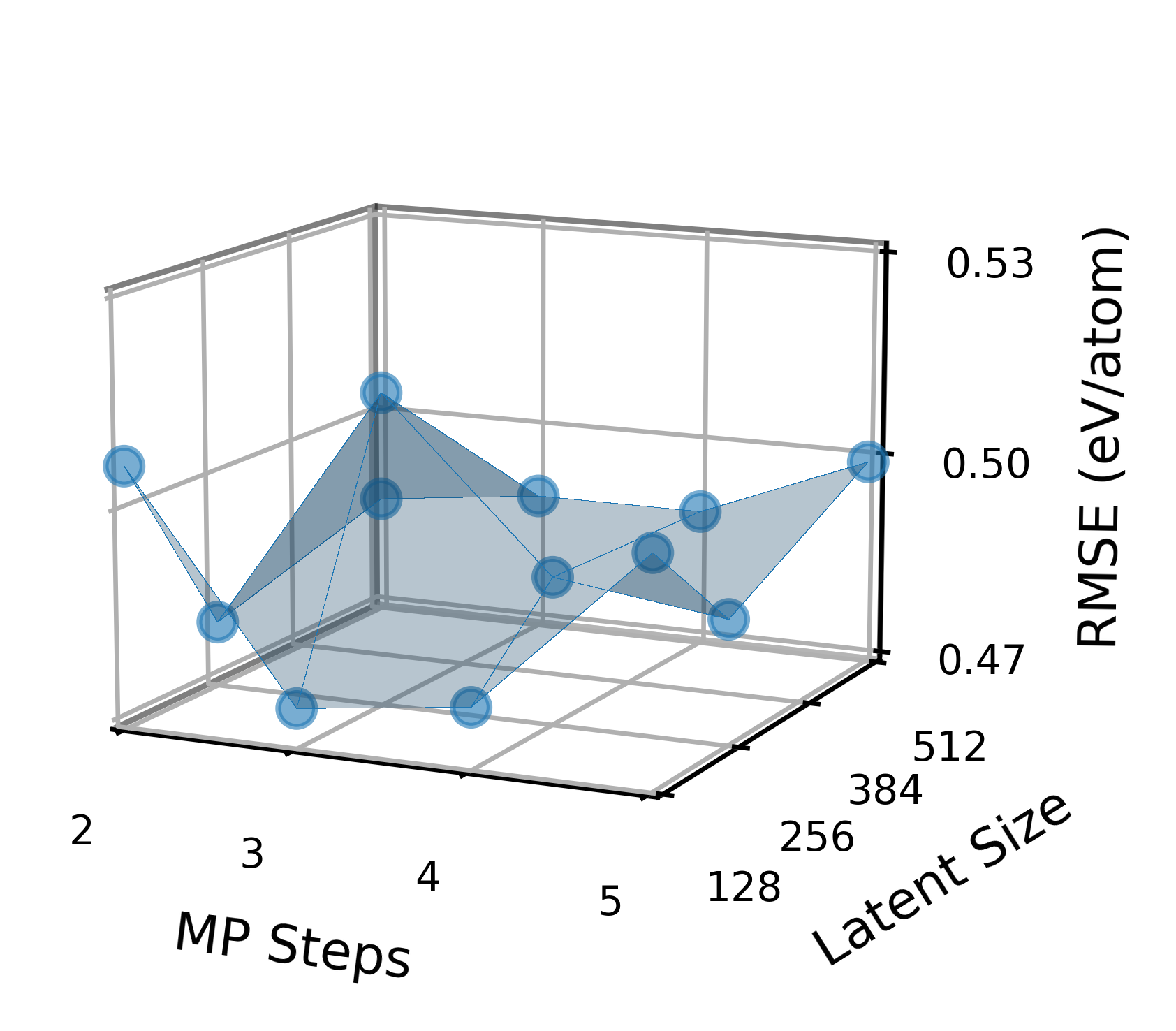}
\caption{Effect of two architecture parameters, number of message passing (MP) steps and latent space size on the validation RMSE of a message passing neural network with edge updates to fit bandgaps of the AFLOW dataset. Other parameters such as the learning rate and size of the readout function were optimized for each MP steps and latent size combination.}
\label{fig:nas_loss_function_3d}
\end{figure}

State of the art in ML is to employ meta-learning approaches, or neural architecture searches (NAS) to find the best NN model architecture~\cite{zoph2016neural}. Typically 500 to 10,000 architectures are selected by an algorithm (\eg random search or reinforcement learning) and trained, and then the best ones are selected by a user-defined reward function~\cite{tan2019mnasnet}. This approach has been applied to a wide variety of fields from object detection~\cite{bender2020can} to audio-signal processing~\cite{speckhard2023neural}, and, more recently, in computational materials science~\cite{bechtel2023band}. It can be formulated mathematically as:
\begin{equation}
\max\limits_{h \in H, \: D \sim P(\mathbf{x})} \mathscr{R}(h(D)),
\label{eq:nas_optimization}
\end{equation}
where the reward function $\mathscr{R}$ is maximized by searching for the NN architecture, $h$, in a user-specified search space, $H$ over some given dataset $D$. The dataset is defined as a collection of independently and identically distributed data points, \ie $D = \{\mathbf{x_1}, \mathbf{x_2},\dots, (\mathbf{x}_n)\}$, where data points $\mathbf{x}_i$ are sampled from some underlying probability distribution $P(\mathbf{x})$. A simple search space may consist of the layers (\eg 1-5) in a deep neural network (DNN) and the nodes per layer (\eg one of [16, 32, 64]). The NAS needs to train each candidate architecture enough to be able to compare them against one another. Note, that there are some methods (\eg differentiable search~\cite{liu2018darts}, shared weights~\cite{bender2020can}, early stopping~\cite{speckhard2023neural}, etc.) in the literature to reduce the computational burden of NAS, but none are guaranteed to return the correct rankings of architectures in terms of the reward function. For an MPNN model, several architecture parameters determine the number of model parameters. Here we discuss two such parameters, the number of message passing steps and the latent space size~\cite{jorgensen2018neural}. There are also other parameters such as the readout function and the number of layers in each graph convolution, which we do not discuss here. Figure \ref{fig:nas_loss_function_3d} shows how the performance of the MPNN architecture described in Ref. ~\citenum{jorgensen2018neural} and trained on AFLOW bandgaps depends on these two parameters. Here, the NAS reward function is the negative of the validation RMSE. Each data point in the figure represents an individual MPNN architecture. The z-axis represents this architecture's best validation RMSE across different initial learning rates, learning decay rate, and readout functions. We see that the validation RMSE as a function of the two architecture parameters is non-convex. This non-convexity is what makes the optimization of the architecture slow and why black-box optimization methods such as reinforcement learning, random search, or evolutionary search are necessary~\cite{real2019regularized}. Whether the NAS reward function is non-convex depends, of course, on the dataset and the search space.

Let us proceed to analyze the computing requirements for a concrete NAS example. The SchNet graph neural network has ca. 85,000 trainable parameters for a latent size of 64 and three message passing steps (when trained on a dataset with 81 different atomic elements)~\cite{schutt2017schnet}. On a single NVIDIA V100 GPU, it takes about two hours to train 2,000,000 steps on the AFLOW bandgap dataset (batch size 32). A further breakdown of the time required per training step, shows that the model takes 0.0010 seconds on average to batch the data and 0.0023 seconds to perform a gradient-update step. For the SchNet-derived NAS architecture search space, approximately 2,000,000 steps are needed to have an idea of whether a candidate architecture is promising or not. This means that to train 2,000 architectures, approximately 40,000 GPU hours for this dataset are needed, or in other words, more than 4.5 years on a single V100 GPU. Converting this number to dollars, current Amazon Web Services pricing plans fluctuate around 3 USD per NVIDIA V100 GPU hour~\cite{aws_ml_pricing}. Thus, a budget of roughly 120,000 USD is needed to perform this NAS. If we use a larger model than SchNet such as PaiNN~\cite{schutt2021equivariant} (600,000 vs 85,000 parameters) the training time and subsequent GPU and dollar budget would increase significantly. Moreover, considering larger datasets than AFLOW (about 60,000 data points, see Section~\ref{sec:transferability_models}) training would also slow down dramatically.

We can also compare CPU versus GPU training for this example. On a single CPU core, the training of 2,000,000 training steps of the SchNet model takes 10.5 hours, or on average approximately 0.0012 seconds each training step to batch the data, and 0.0177 seconds to take a gradient step and update the parameters. We conclude that CPU training is nearly an order of magnitude slower.

As NAS becomes more ubiquitous in materials science, as it has in other fields, new infrastructure challenges will have to be addressed. Finding the optimal model is often a non-convex task and can belong to large multidimensional search spaces that requires black-box optimization methods. For models with a relatively small number of parameters, an MPNN NAS can require a huge computational and therefore monetary budget to be run on medium-sized to large datasets. As datasets and the base models being trained grow larger, so too will the infrastructure requirements.

\section{Discussion and conclusions}

If we define a dataset as "big" because it has a large number of data points, we may wonder why ML models fail to generalize. For this aspect, data diversity, \ie how well the underlying physical space is sampled, is critical. For example, AFLOW and MP datasets, despite their size, do not appear to contain enough data to describe the wide variety of materials. In other words, even larger and --in particular-- more diverse datasets are required to build robust and transferable models. This has been shown by combining these two datasets.

Efforts to create {\it big} datasets will require advanced technical and software solutions. For example, computing only a fraction of above mentioned datasets, causes file and storage issues. In addition, as datasets grow, so does the number of model parameters. MPNN and other variants of graph neural networks have become quite common in materials science. Finding the best MPNN model, requires a neural-architecture search over an often non-convex search space. GPUs offer an order of magnitude speedup in training, but using enough GPUs to perform a single NAS search over a large dataset requires a significant monetary budget. 

Databases also pose challenges related to data veracity. For instance, data points computed for the same material, may differ in the input settings. Sorting or clustering data using similarity metrics can help users better understand the quality of data. It also enables ML practitioners to filter data into homogeneous subsets that contain less variation in the target properties. Multi-fidelity modeling~\cite{huang2020quantum}, where hetereogenous data can be used by the model, is not discussed here, would be an alternative option for dealing with data veracity.

Big data presents an opportunity to use complex models. Complexity is shown to be well defined within the confines of a single model type. A general definition for comparative purposes is, however, lacking. The total trainable parameter count, or the number of constant, linear, or nonlinear terms can be less important than the model class when predicting performance. Increasing complexity, however, as shown with the example of using nonlinear features in cluster expansion, can aide significantly in describing intricate physics. More research is needed to better define model complexity in a data- and model- independent manner.

In this work, we attempt to shine light on some aspects of Big Data in materials science. There are certainly many more aspects to be explored. We hope that the issues we illustrate here will motivate further research along these lines.

\section*{Code availability}
The code used to gather and generate AFLOW dataset splits, perform neural architecture searches on MPNNs and profile the batching and gradient update steps of MPNNs can be found here: \url{https://github.com/tisabe/jraph_mpeu}. The SISSO and NN models fit to Ref.~\citenum{carbogno2022numerical} can be found at this link: \url{https://nomad-lab.eu/aitutorials/error-estimates-qrf}. The NN and RF models fit to the AFLOW bandgaps can be found here \url{https://colab.research.google.com/drive/1d52Z5QLC9qdU06xTa6X9c1UGEylcrkoq?usp=sharing}.

\section*{Author Contributions}
C.D. sketched the idea and drafted the manuscript. T.B. and D.S. provided the example on model transferability; S.R. provided the cluster-expansion example. M.K. provided the example on similarity measures. D.S. and L.G. led the discussion on model complexity. D.S. and T.B. provided the example on infrastructure requirements. All authors contributed to the writing of the manuscript.

\section*{Conflicts of interest}
There are no conflicts to declare.

\section*{Acknowledgements}
Work carried out in part by the Max Planck Graduate Center for Quantum Materials. Partial funding is appreciated from the German Science Foundation (DFG) through the CRC FONDA (project 414984028) and the NIFD consortium FAIRmat (project 460197019), and the European Union’s Horizon 2020 research and innovation program under the grant agreement Nº 951786 (NOMAD CoE). D.S. acknowledges support by the IMPRS for Elementary Processes in Physical Chemistry. Computing time for the \exciting\ workflow example was granted by the Resource Allocation Board and provided on the supercomputers Lise and Emmy at NHR@ZIB and NHR@Göttingen as part of the NHR infrastructure (bep00098). We thank Sebastian Kehl at the Max Planck Data Computing Facility for his advice on taking CPU/GPU profiling measurements.

\bibliography{main.bib} 

\providecommand*{\mcitethebibliography}{\thebibliography}
\csname @ifundefined\endcsname{endmcitethebibliography}
{\let\endmcitethebibliography\endthebibliography}{}
\begin{mcitethebibliography}{46}
\providecommand*{\natexlab}[1]{#1}
\providecommand*{\mciteSetBstSublistMode}[1]{}
\providecommand*{\mciteSetBstMaxWidthForm}[2]{}
\providecommand*{\mciteBstWouldAddEndPuncttrue}
  {\def\EndOfBibitem{\unskip.}}
\providecommand*{\mciteBstWouldAddEndPunctfalse}
  {\let\EndOfBibitem\relax}
\providecommand*{\mciteSetBstMidEndSepPunct}[3]{}
\providecommand*{\mciteSetBstSublistLabelBeginEnd}[3]{}
\providecommand*{\EndOfBibitem}{}
\mciteSetBstSublistMode{f}
\mciteSetBstMaxWidthForm{subitem}
{(\emph{\alph{mcitesubitemcount}})}
\mciteSetBstSublistLabelBeginEnd{\mcitemaxwidthsubitemform\space}
{\relax}{\relax}

\bibitem[Draxl and Scheffler(2020)]{draxl2020big}
C.~Draxl and M.~Scheffler, \emph{Handbook of Materials Modeling: Methods:
  Theory and Modeling}, 2020,  49--73\relax
\mciteBstWouldAddEndPuncttrue
\mciteSetBstMidEndSepPunct{\mcitedefaultmidpunct}
{\mcitedefaultendpunct}{\mcitedefaultseppunct}\relax
\EndOfBibitem
\bibitem[Speckhard \emph{et~al.}(2023)Speckhard, Carbogno, Ghiringhelli,
  Lubeck, Scheffler, and Draxl]{speckhard2023extrapolation}
D.~T. Speckhard, C.~Carbogno, L.~Ghiringhelli, S.~Lubeck, M.~Scheffler and
  C.~Draxl, \emph{arXiv preprint arXiv:2303.14760}, 2023\relax
\mciteBstWouldAddEndPuncttrue
\mciteSetBstMidEndSepPunct{\mcitedefaultmidpunct}
{\mcitedefaultendpunct}{\mcitedefaultseppunct}\relax
\EndOfBibitem
\bibitem[Jha \emph{et~al.}(2018)Jha, Ward, Paul, Liao, Choudhary, Wolverton,
  and Agrawal]{jha2018elemnet}
D.~Jha, L.~Ward, A.~Paul, W.-k. Liao, A.~Choudhary, C.~Wolverton and
  A.~Agrawal, \emph{Scientific reports}, 2018, \textbf{8}, 17593\relax
\mciteBstWouldAddEndPuncttrue
\mciteSetBstMidEndSepPunct{\mcitedefaultmidpunct}
{\mcitedefaultendpunct}{\mcitedefaultseppunct}\relax
\EndOfBibitem
\bibitem[Jain \emph{et~al.}(2013)Jain, Ong, Hautier, Chen, Richards, Dacek,
  Cholia, Gunter, Skinner, Ceder, and Persson]{MP}
A.~Jain, S.~P. Ong, G.~Hautier, W.~Chen, W.~D. Richards, S.~Dacek, S.~Cholia,
  D.~Gunter, D.~Skinner, G.~Ceder and K.~A. Persson, \emph{APL Materials},
  2013, \textbf{1}, 011002\relax
\mciteBstWouldAddEndPuncttrue
\mciteSetBstMidEndSepPunct{\mcitedefaultmidpunct}
{\mcitedefaultendpunct}{\mcitedefaultseppunct}\relax
\EndOfBibitem
\bibitem[Calderon \emph{et~al.}(2015)Calderon, Plata, Toher, Oses, Levy,
  Fornari, Natan, Mehl, Hart, Nardelli,\emph{et~al.}]{AFLOW}
C.~E. Calderon, J.~J. Plata, C.~Toher, C.~Oses, O.~Levy, M.~Fornari, A.~Natan,
  M.~J. Mehl, G.~Hart, M.~B. Nardelli \emph{et~al.}, \emph{Computational
  Materials Science}, 2015, \textbf{108}, 233--238\relax
\mciteBstWouldAddEndPuncttrue
\mciteSetBstMidEndSepPunct{\mcitedefaultmidpunct}
{\mcitedefaultendpunct}{\mcitedefaultseppunct}\relax
\EndOfBibitem
\bibitem[Bechtel \emph{et~al.}(2023)Bechtel, Speckhard, Godwin, and
  Draxl]{bechtel2023band}
T.~Bechtel, D.~T. Speckhard, J.~Godwin and C.~Draxl, \emph{arXiv:2309.06348},
  2023\relax
\mciteBstWouldAddEndPuncttrue
\mciteSetBstMidEndSepPunct{\mcitedefaultmidpunct}
{\mcitedefaultendpunct}{\mcitedefaultseppunct}\relax
\EndOfBibitem
\bibitem[Chen \emph{et~al.}(2019)Chen, Ye, Zuo, Zheng, and Ong]{chen2019graph}
C.~Chen, W.~Ye, Y.~Zuo, C.~Zheng and S.~P. Ong, \emph{Chemistry of Materials},
  2019, \textbf{31}, 3564--3572\relax
\mciteBstWouldAddEndPuncttrue
\mciteSetBstMidEndSepPunct{\mcitedefaultmidpunct}
{\mcitedefaultendpunct}{\mcitedefaultseppunct}\relax
\EndOfBibitem
\bibitem[Schmidt \emph{et~al.}(2021)Schmidt, Pettersson, Verdozzi, Botti, and
  Marques]{schmidt2021crystal}
J.~Schmidt, L.~Pettersson, C.~Verdozzi, S.~Botti and M.~A. Marques,
  \emph{Science advances}, 2021, \textbf{7}, eabi7948\relax
\mciteBstWouldAddEndPuncttrue
\mciteSetBstMidEndSepPunct{\mcitedefaultmidpunct}
{\mcitedefaultendpunct}{\mcitedefaultseppunct}\relax
\EndOfBibitem
\bibitem[J{\o}rgensen \emph{et~al.}(2018)J{\o}rgensen, Jacobsen, and
  Schmidt]{jorgensen2018neural}
P.~B. J{\o}rgensen, K.~W. Jacobsen and M.~N. Schmidt, \emph{arXiv preprint
  arXiv:1806.03146}, 2018\relax
\mciteBstWouldAddEndPuncttrue
\mciteSetBstMidEndSepPunct{\mcitedefaultmidpunct}
{\mcitedefaultendpunct}{\mcitedefaultseppunct}\relax
\EndOfBibitem
\bibitem[Lejaeghere \emph{et~al.}(2016)Lejaeghere, Bihlmayer, Bj{\"o}rkman,
  Blaha, Bl{\"u}gel, Blum, Caliste, Castelli, Clark,
  Dal~Corso,\emph{et~al.}]{lejaeghere2016reproducibility}
K.~Lejaeghere, G.~Bihlmayer, T.~Bj{\"o}rkman, P.~Blaha, S.~Bl{\"u}gel, V.~Blum,
  D.~Caliste, I.~E. Castelli, S.~J. Clark, A.~Dal~Corso \emph{et~al.},
  \emph{Science}, 2016, \textbf{351}, aad3000\relax
\mciteBstWouldAddEndPuncttrue
\mciteSetBstMidEndSepPunct{\mcitedefaultmidpunct}
{\mcitedefaultendpunct}{\mcitedefaultseppunct}\relax
\EndOfBibitem
\bibitem[Huang \emph{et~al.}(2020)Huang, Symonds, and von
  Lilienfeld]{huang2020quantum}
B.~Huang, N.~O. Symonds and O.~A. von Lilienfeld, \emph{Handbook of Materials
  Modeling: Methods: Theory and Modeling}, 2020,  1883--1909\relax
\mciteBstWouldAddEndPuncttrue
\mciteSetBstMidEndSepPunct{\mcitedefaultmidpunct}
{\mcitedefaultendpunct}{\mcitedefaultseppunct}\relax
\EndOfBibitem
\bibitem[Kuban \emph{et~al.}(2022)Kuban, Gabaj, Aggoune, Vona, Rigamonti, and
  Draxl]{Kuban2022MRS}
M.~Kuban, {\v{S}}.~Gabaj, W.~Aggoune, C.~Vona, S.~Rigamonti and C.~Draxl,
  \emph{MRS Bulletin}, 2022, \textbf{47}, 991--999\relax
\mciteBstWouldAddEndPuncttrue
\mciteSetBstMidEndSepPunct{\mcitedefaultmidpunct}
{\mcitedefaultendpunct}{\mcitedefaultseppunct}\relax
\EndOfBibitem
\bibitem[Kuban \emph{et~al.}(2024)Kuban, Rigamonti, and Draxl]{Kuban2024MADAS}
M.~Kuban, S.~Rigamonti and C.~Draxl, \emph{MADAS -- A Python framework for
  assessing similarity in materials-science data}, 2024\relax
\mciteBstWouldAddEndPuncttrue
\mciteSetBstMidEndSepPunct{\mcitedefaultmidpunct}
{\mcitedefaultendpunct}{\mcitedefaultseppunct}\relax
\EndOfBibitem
\bibitem[Draxl and Scheffler(2018)]{NOMAD}
C.~Draxl and M.~Scheffler, \emph{Mrs Bulletin}, 2018, \textbf{43},
  676--682\relax
\mciteBstWouldAddEndPuncttrue
\mciteSetBstMidEndSepPunct{\mcitedefaultmidpunct}
{\mcitedefaultendpunct}{\mcitedefaultseppunct}\relax
\EndOfBibitem
\bibitem[NOMAD(2022)]{CarbognoFHIaimsErrorsData}
NOMAD, \emph{Numerical Errors FHI-aims Dataset}, 2022,
  \url{https://dx.doi.org/10.17172/NOMAD/2020.07.27-1}\relax
\mciteBstWouldAddEndPuncttrue
\mciteSetBstMidEndSepPunct{\mcitedefaultmidpunct}
{\mcitedefaultendpunct}{\mcitedefaultseppunct}\relax
\EndOfBibitem
\bibitem[Blum \emph{et~al.}(2009)Blum, Gehrke, Hanke, Havu, Havu, Ren, Reuter,
  and Scheffler]{Blum2009FHIaims}
V.~Blum, R.~Gehrke, F.~Hanke, P.~Havu, V.~Havu, X.~Ren, K.~Reuter and
  M.~Scheffler, \emph{Computer Physics Communications}, 2009, \textbf{180},
  2175--2196\relax
\mciteBstWouldAddEndPuncttrue
\mciteSetBstMidEndSepPunct{\mcitedefaultmidpunct}
{\mcitedefaultendpunct}{\mcitedefaultseppunct}\relax
\EndOfBibitem
\bibitem[Carbogno \emph{et~al.}(2022)Carbogno, Thygesen, Bieniek, Draxl,
  Ghiringhelli, Gulans, Hofmann, Jacobsen, Lubeck,
  Mortensen,\emph{et~al.}]{carbogno2022numerical}
C.~Carbogno, K.~S. Thygesen, B.~Bieniek, C.~Draxl, L.~M. Ghiringhelli,
  A.~Gulans, O.~T. Hofmann, K.~W. Jacobsen, S.~Lubeck, J.~J. Mortensen
  \emph{et~al.}, \emph{npj Computational Materials}, 2022, \textbf{8}, 69\relax
\mciteBstWouldAddEndPuncttrue
\mciteSetBstMidEndSepPunct{\mcitedefaultmidpunct}
{\mcitedefaultendpunct}{\mcitedefaultseppunct}\relax
\EndOfBibitem
\bibitem[Kuban \emph{et~al.}(2022)Kuban, Rigamonti, Scheidgen, and
  Draxl]{Kuban2022SDATA}
M.~Kuban, S.~Rigamonti, M.~Scheidgen and C.~Draxl, \emph{Scientific Data},
  2022, \textbf{9}, 646\relax
\mciteBstWouldAddEndPuncttrue
\mciteSetBstMidEndSepPunct{\mcitedefaultmidpunct}
{\mcitedefaultendpunct}{\mcitedefaultseppunct}\relax
\EndOfBibitem
\bibitem[Sanchez \emph{et~al.}(1984)Sanchez, Ducastelle, and
  Gratias]{Sanchez1984}
J.~Sanchez, F.~Ducastelle and D.~Gratias, \emph{Physica A: Statistical
  Mechanics and its Applications}, 1984, \textbf{128}, 334 -- 350\relax
\mciteBstWouldAddEndPuncttrue
\mciteSetBstMidEndSepPunct{\mcitedefaultmidpunct}
{\mcitedefaultendpunct}{\mcitedefaultseppunct}\relax
\EndOfBibitem
\bibitem[Rigamonti \emph{et~al.}(2023)Rigamonti, Troppenz, Kuban, Huebner, and
  Draxl]{rigamonti2023cell}
S.~Rigamonti, M.~Troppenz, M.~Kuban, A.~Huebner and C.~Draxl,
  \emph{arXiv:2310.18223}, 2023\relax
\mciteBstWouldAddEndPuncttrue
\mciteSetBstMidEndSepPunct{\mcitedefaultmidpunct}
{\mcitedefaultendpunct}{\mcitedefaultseppunct}\relax
\EndOfBibitem
\bibitem[Stroth \emph{et~al.}(2024)Stroth, Rigamonti, and
  Draxl]{stroth2024nonlinearce}
A.~Stroth, S.~Rigamonti and C.~Draxl, \emph{preprint}, 2024\relax
\mciteBstWouldAddEndPuncttrue
\mciteSetBstMidEndSepPunct{\mcitedefaultmidpunct}
{\mcitedefaultendpunct}{\mcitedefaultseppunct}\relax
\EndOfBibitem
\bibitem[Ouyang \emph{et~al.}(2018)Ouyang, Curtarolo, Ahmetcik, Scheffler, and
  Ghiringhelli]{ouyang2018sisso}
R.~Ouyang, S.~Curtarolo, E.~Ahmetcik, M.~Scheffler and L.~M. Ghiringhelli,
  \emph{Physical Review Materials}, 2018, \textbf{2}, 083802\relax
\mciteBstWouldAddEndPuncttrue
\mciteSetBstMidEndSepPunct{\mcitedefaultmidpunct}
{\mcitedefaultendpunct}{\mcitedefaultseppunct}\relax
\EndOfBibitem
\bibitem[Bianchini and Scarselli(2014)]{nn_complexity}
M.~Bianchini and F.~Scarselli, \emph{IEEE Transactions on Neural Networks and
  Learning Systems}, 2014, \textbf{25}, 1553--1565\relax
\mciteBstWouldAddEndPuncttrue
\mciteSetBstMidEndSepPunct{\mcitedefaultmidpunct}
{\mcitedefaultendpunct}{\mcitedefaultseppunct}\relax
\EndOfBibitem
\bibitem[Purcell \emph{et~al.}(2023)Purcell, Scheffler, and
  Ghiringhelli]{purcell2023recent}
T.~A.~R. Purcell, M.~Scheffler and L.~M. Ghiringhelli, \emph{The Journal of
  Chemical Physics}, 2023, \textbf{159}, 114110\relax
\mciteBstWouldAddEndPuncttrue
\mciteSetBstMidEndSepPunct{\mcitedefaultmidpunct}
{\mcitedefaultendpunct}{\mcitedefaultseppunct}\relax
\EndOfBibitem
\bibitem[Aggarwal \emph{et~al.}(2018)Aggarwal\emph{et~al.}]{aggarwal2018neural}
C.~C. Aggarwal \emph{et~al.}, \emph{Neural networks and deep learning},
  Springer, 2018, vol.~10\relax
\mciteBstWouldAddEndPuncttrue
\mciteSetBstMidEndSepPunct{\mcitedefaultmidpunct}
{\mcitedefaultendpunct}{\mcitedefaultseppunct}\relax
\EndOfBibitem
\bibitem[Ward \emph{et~al.}(2016)Ward, Agrawal, Choudhary, and
  Wolverton]{ward2016general}
L.~Ward, A.~Agrawal, A.~Choudhary and C.~Wolverton, \emph{npj Computational
  Materials}, 2016, \textbf{2}, 1--7\relax
\mciteBstWouldAddEndPuncttrue
\mciteSetBstMidEndSepPunct{\mcitedefaultmidpunct}
{\mcitedefaultendpunct}{\mcitedefaultseppunct}\relax
\EndOfBibitem
\bibitem[Rosen \emph{et~al.}(2024)Rosen, Gallant, George, Riebesell,
  Sahasrabuddhe, Shen, Wen, Evans, Petretto,
  Waroquiers,\emph{et~al.}]{rosen2024jobflow}
A.~S. Rosen, M.~Gallant, J.~George, J.~Riebesell, H.~Sahasrabuddhe, J.-X. Shen,
  M.~Wen, M.~L. Evans, G.~Petretto, D.~Waroquiers \emph{et~al.}, \emph{Journal
  of Open Source Software}, 2024, \textbf{9}, 5995\relax
\mciteBstWouldAddEndPuncttrue
\mciteSetBstMidEndSepPunct{\mcitedefaultmidpunct}
{\mcitedefaultendpunct}{\mcitedefaultseppunct}\relax
\EndOfBibitem
\bibitem[Schintke \emph{et~al.}(2024)Schintke, Belhajjame, De~Mecquenem,
  Frantz, Guarino, Hilbrich, Lehmann, Missier, Sattler,
  Sparka,\emph{et~al.}]{schintke2024validity}
F.~Schintke, K.~Belhajjame, N.~De~Mecquenem, D.~Frantz, V.~E. Guarino,
  M.~Hilbrich, F.~Lehmann, P.~Missier, R.~Sattler, J.~A. Sparka \emph{et~al.},
  \emph{Future Generation Computer Systems}, 2024, \textbf{157}, 82--97\relax
\mciteBstWouldAddEndPuncttrue
\mciteSetBstMidEndSepPunct{\mcitedefaultmidpunct}
{\mcitedefaultendpunct}{\mcitedefaultseppunct}\relax
\EndOfBibitem
\bibitem[Larsen \emph{et~al.}(2017)Larsen, Mortensen, Blomqvist, Castelli,
  Christensen, Du{\l}ak, Friis, Groves, Hammer,
  Hargus,\emph{et~al.}]{larsen2017atomic}
A.~H. Larsen, J.~J. Mortensen, J.~Blomqvist, I.~E. Castelli, R.~Christensen,
  M.~Du{\l}ak, J.~Friis, M.~N. Groves, B.~Hammer, C.~Hargus \emph{et~al.},
  \emph{Journal of Physics: Condensed Matter}, 2017, \textbf{29}, 273002\relax
\mciteBstWouldAddEndPuncttrue
\mciteSetBstMidEndSepPunct{\mcitedefaultmidpunct}
{\mcitedefaultendpunct}{\mcitedefaultseppunct}\relax
\EndOfBibitem
\bibitem[Gulans \emph{et~al.}(2014)Gulans, Kontur, Meisenbichler, Nabok,
  Pavone, Rigamonti, Sagmeister, Werner, and Draxl]{gulans2014exciting}
A.~Gulans, S.~Kontur, C.~Meisenbichler, D.~Nabok, P.~Pavone, S.~Rigamonti,
  S.~Sagmeister, U.~Werner and C.~Draxl, \emph{Journal of Physics: Condensed
  Matter}, 2014, \textbf{26}, 363202\relax
\mciteBstWouldAddEndPuncttrue
\mciteSetBstMidEndSepPunct{\mcitedefaultmidpunct}
{\mcitedefaultendpunct}{\mcitedefaultseppunct}\relax
\EndOfBibitem
\bibitem[Buccheri \emph{et~al.}(2023)Buccheri, Peschel, Maurer, Voiculescu,
  Speckhard, Kleine, Stephan, Kuban, and Draxl]{buccheri2023excitingtools}
A.~Buccheri, F.~Peschel, B.~Maurer, M.~Voiculescu, D.~T. Speckhard, H.~Kleine,
  E.~Stephan, M.~Kuban and C.~Draxl, \emph{Journal of Open Source Software},
  2023, \textbf{8}, 5148\relax
\mciteBstWouldAddEndPuncttrue
\mciteSetBstMidEndSepPunct{\mcitedefaultmidpunct}
{\mcitedefaultendpunct}{\mcitedefaultseppunct}\relax
\EndOfBibitem
\bibitem[NOMAD(2024)]{nomad_oasis_webpage}
NOMAD, \emph{Nomad Oasis Webpage}, 2024,
  \url{https://nomad-lab.eu/nomad-lab/nomad-oasis.html}\relax
\mciteBstWouldAddEndPuncttrue
\mciteSetBstMidEndSepPunct{\mcitedefaultmidpunct}
{\mcitedefaultendpunct}{\mcitedefaultseppunct}\relax
\EndOfBibitem
\bibitem[Scheffler \emph{et~al.}(2022)Scheffler, Aeschlimann, Albrecht, Bereau,
  Bungartz, Felser, Greiner, Gro{\ss}, Koch,
  Kremer,\emph{et~al.}]{scheffler2022fair}
M.~Scheffler, M.~Aeschlimann, M.~Albrecht, T.~Bereau, H.-J. Bungartz,
  C.~Felser, M.~Greiner, A.~Gro{\ss}, C.~T. Koch, K.~Kremer \emph{et~al.},
  \emph{Nature}, 2022, \textbf{604}, 635--642\relax
\mciteBstWouldAddEndPuncttrue
\mciteSetBstMidEndSepPunct{\mcitedefaultmidpunct}
{\mcitedefaultendpunct}{\mcitedefaultseppunct}\relax
\EndOfBibitem
\bibitem[Scheidgen \emph{et~al.}(2023)Scheidgen, Himanen, Ladines, Sikter,
  Nakhaee, Ádám Fekete, Chang, Golparvar, Márquez, Brockhauser, Brückner,
  Ghiringhelli, Dietrich, Lehmberg, Denell, Albino, Näsström, Shabih,
  Dobener, Kühbach, Mozumder, Rudzinski, Daelman, Pizarro, Kuban, Salazar,
  Ondračka, Bungartz, and Draxl]{Scheidgen2023}
M.~Scheidgen, L.~Himanen, A.~N. Ladines, D.~Sikter, M.~Nakhaee, Ádám Fekete,
  T.~Chang, A.~Golparvar, J.~A. Márquez, S.~Brockhauser, S.~Brückner, L.~M.
  Ghiringhelli, F.~Dietrich, D.~Lehmberg, T.~Denell, A.~Albino, H.~Näsström,
  S.~Shabih, F.~Dobener, M.~Kühbach, R.~Mozumder, J.~F. Rudzinski, N.~Daelman,
  J.~M. Pizarro, M.~Kuban, C.~Salazar, P.~Ondračka, H.-J. Bungartz and
  C.~Draxl, \emph{Journal of Open Source Software}, 2023, \textbf{8},
  5388\relax
\mciteBstWouldAddEndPuncttrue
\mciteSetBstMidEndSepPunct{\mcitedefaultmidpunct}
{\mcitedefaultendpunct}{\mcitedefaultseppunct}\relax
\EndOfBibitem
\bibitem[Sbail{\`o} \emph{et~al.}(2022)Sbail{\`o}, Fekete, Ghiringhelli, and
  Scheffler]{sbailo2022nomad}
L.~Sbail{\`o}, {\'A}.~Fekete, L.~M. Ghiringhelli and M.~Scheffler, \emph{npj
  Computational Materials}, 2022, \textbf{8}, 250\relax
\mciteBstWouldAddEndPuncttrue
\mciteSetBstMidEndSepPunct{\mcitedefaultmidpunct}
{\mcitedefaultendpunct}{\mcitedefaultseppunct}\relax
\EndOfBibitem
\bibitem[Draxl and Scheffler(2019)]{draxl2019nomad}
C.~Draxl and M.~Scheffler, \emph{Journal of Physics: Materials}, 2019,
  \textbf{2}, 036001\relax
\mciteBstWouldAddEndPuncttrue
\mciteSetBstMidEndSepPunct{\mcitedefaultmidpunct}
{\mcitedefaultendpunct}{\mcitedefaultseppunct}\relax
\EndOfBibitem
\bibitem[Ghiringhelli \emph{et~al.}(2023)Ghiringhelli, Baldauf, Bereau,
  Brockhauser, Carbogno, Chamanara, Cozzini, Curtarolo, Draxl,
  Dwaraknath,\emph{et~al.}]{ghiringhelli2023shared}
L.~M. Ghiringhelli, C.~Baldauf, T.~Bereau, S.~Brockhauser, C.~Carbogno,
  J.~Chamanara, S.~Cozzini, S.~Curtarolo, C.~Draxl, S.~Dwaraknath
  \emph{et~al.}, \emph{Scientific Data}, 2023, \textbf{10}, 626\relax
\mciteBstWouldAddEndPuncttrue
\mciteSetBstMidEndSepPunct{\mcitedefaultmidpunct}
{\mcitedefaultendpunct}{\mcitedefaultseppunct}\relax
\EndOfBibitem
\bibitem[Zoph and Le(2016)]{zoph2016neural}
B.~Zoph and Q.~V. Le, \emph{arXiv preprint arXiv:1611.01578}, 2016\relax
\mciteBstWouldAddEndPuncttrue
\mciteSetBstMidEndSepPunct{\mcitedefaultmidpunct}
{\mcitedefaultendpunct}{\mcitedefaultseppunct}\relax
\EndOfBibitem
\bibitem[Tan \emph{et~al.}(2019)Tan, Chen, Pang, Vasudevan, Sandler, Howard,
  and Le]{tan2019mnasnet}
M.~Tan, B.~Chen, R.~Pang, V.~Vasudevan, M.~Sandler, A.~Howard and Q.~V. Le,
  Proceedings of the IEEE/CVF conference on computer vision and pattern
  recognition, 2019, pp. 2820--2828\relax
\mciteBstWouldAddEndPuncttrue
\mciteSetBstMidEndSepPunct{\mcitedefaultmidpunct}
{\mcitedefaultendpunct}{\mcitedefaultseppunct}\relax
\EndOfBibitem
\bibitem[Bender \emph{et~al.}(2020)Bender, Liu, Chen, Chu, Cheng, Kindermans,
  and Le]{bender2020can}
G.~Bender, H.~Liu, B.~Chen, G.~Chu, S.~Cheng, P.-J. Kindermans and Q.~V. Le,
  Proceedings of the IEEE/CVF conference on computer vision and pattern
  recognition, 2020, pp. 14323--14332\relax
\mciteBstWouldAddEndPuncttrue
\mciteSetBstMidEndSepPunct{\mcitedefaultmidpunct}
{\mcitedefaultendpunct}{\mcitedefaultseppunct}\relax
\EndOfBibitem
\bibitem[Speckhard \emph{et~al.}(2023)Speckhard, Misiunas, Perel, Zhu, Carlile,
  and Slaney]{speckhard2023neural}
D.~T. Speckhard, K.~Misiunas, S.~Perel, T.~Zhu, S.~Carlile and M.~Slaney,
  \emph{Neural Computing and Applications}, 2023, \textbf{35},
  12133--12144\relax
\mciteBstWouldAddEndPuncttrue
\mciteSetBstMidEndSepPunct{\mcitedefaultmidpunct}
{\mcitedefaultendpunct}{\mcitedefaultseppunct}\relax
\EndOfBibitem
\bibitem[Liu \emph{et~al.}(2018)Liu, Simonyan, and Yang]{liu2018darts}
H.~Liu, K.~Simonyan and Y.~Yang, \emph{arXiv preprint arXiv:1806.09055},
  2018\relax
\mciteBstWouldAddEndPuncttrue
\mciteSetBstMidEndSepPunct{\mcitedefaultmidpunct}
{\mcitedefaultendpunct}{\mcitedefaultseppunct}\relax
\EndOfBibitem
\bibitem[Real \emph{et~al.}(2019)Real, Aggarwal, Huang, and
  Le]{real2019regularized}
E.~Real, A.~Aggarwal, Y.~Huang and Q.~V. Le, Proceedings of the aaai conference
  on artificial intelligence, 2019, pp. 4780--4789\relax
\mciteBstWouldAddEndPuncttrue
\mciteSetBstMidEndSepPunct{\mcitedefaultmidpunct}
{\mcitedefaultendpunct}{\mcitedefaultseppunct}\relax
\EndOfBibitem
\bibitem[Sch\"{u}tt \emph{et~al.}(2017)Sch\"{u}tt, Kindermans, Sauceda~Felix,
  Chmiela, Tkatchenko, and M\"{u}ller]{schutt2017schnet}
K.~Sch\"{u}tt, P.-J. Kindermans, H.~E. Sauceda~Felix, S.~Chmiela, A.~Tkatchenko
  and K.-R. M\"{u}ller, Advances in Neural Information Processing Systems,
  2017\relax
\mciteBstWouldAddEndPuncttrue
\mciteSetBstMidEndSepPunct{\mcitedefaultmidpunct}
{\mcitedefaultendpunct}{\mcitedefaultseppunct}\relax
\EndOfBibitem
\bibitem[Amazon(2024)]{aws_ml_pricing}
Amazon, \emph{Amazon Web Services EC2 P3 Instances}, 2024,
  \url{https://aws.amazon.com/ec2/instance-types/p3/}\relax
\mciteBstWouldAddEndPuncttrue
\mciteSetBstMidEndSepPunct{\mcitedefaultmidpunct}
{\mcitedefaultendpunct}{\mcitedefaultseppunct}\relax
\EndOfBibitem
\bibitem[Sch{\"u}tt \emph{et~al.}(2021)Sch{\"u}tt, Unke, and
  Gastegger]{schutt2021equivariant}
K.~Sch{\"u}tt, O.~Unke and M.~Gastegger, International Conference on Machine
  Learning, 2021, pp. 9377--9388\relax
\mciteBstWouldAddEndPuncttrue
\mciteSetBstMidEndSepPunct{\mcitedefaultmidpunct}
{\mcitedefaultendpunct}{\mcitedefaultseppunct}\relax
\EndOfBibitem
\end{mcitethebibliography}
\bibliographystyle{rsc} 

\end{document}